\documentclass[onecolumn,a4paper,conference]{IEEEtran}

\usepackage{cite}
\usepackage{amsmath,amssymb,amsfonts}
\usepackage{algorithmic}
\usepackage{graphicx}
\usepackage{textcomp}
\usepackage{xcolor}

\usepackage{subfigure}
\usepackage{dsfont}
\usepackage{multirow}
\usepackage[ruled, norelsize]{algorithm2e}
\usepackage{cite}

\ifCLASSINFOpdf
\else
\fi
\def\BibTeX{{\rm B\kern-.05em{\sc i\kern-.025em b}\kern-.08em
    T\kern-.1667em\lower.7ex\hbox{E}\kern-.125emX}}
\begin{document}
%
\title{Leachable Component Clustering	\\
\thanks{\textsuperscript{*}Corresponding author.}
}


\author{\IEEEauthorblockN{Miao Cheng\textsuperscript{*}
}
\IEEEauthorblockA{School of Computer Science \\ and Engineering\\
Guangxi Normal University	\\
Guilin, Guangxi, China	\\
Email: mcheng@mailbox.gxnu.edu.cn}
\and
\IEEEauthorblockN{Xinge You}
\IEEEauthorblockA{School of Electronic Information 	\\
and Communication	\\
Huazhong University of Science and Technology	\\
Wuhan, Hubei, China	\\
Email: youxg@hust.edu.cn}
}


%


\maketitle

\begin{abstract}
Clustering attempts to partition data instances into several distinctive groups, while the similarities among data belonging to the common partition can be principally reserved. Furthermore, incomplete data frequently occurs in many real-world applications, and brings perverse influence on pattern analysis. As a consequence, the specific solutions to data imputation and handling are developed to conduct the missing values of data, and independent stage of knowledge exploitation is absorbed for information understanding. In this work, a novel approach to clustering of incomplete data, termed leachable component clustering, is proposed. Rather than existing methods, the proposed method handles data imputation with Bayes alignment, and collects the lost patterns in theory. Due to the simple numeric computation of equations, the proposed method can learn optimized partitions while the calculation efficiency is held. Experiments on several artificial incomplete data sets demonstrate that, the proposed method is able to present superior performance compared with other state-of-the-art algorithms.
\end{abstract}


\section{Introduction}
As an essential category of data analysis and handling methods, clustering aims to divide data instances into several separate data groups, by assuming the instances of each cluster share the common similarities of data patterns \cite{Duda00PC}\cite{Bishop11PRML}. As a consequence, different data points are to be organized regularly in accordance to certain partition rules, and normally, further handling can be achieved conveniently \cite{Vidal11SC}. 
Until now, \emph{k}-means has been the most popular solution to most clustering problems, due to its stable and outstanding performance \cite{Kanungo02KMeans}, which is derived from the previous contribution of the classic Lloyd's method \cite{Lloyd82LSQ}. Furthermore, spectral clustering method \cite{Ng01SC}\cite{Manor04STSC} has received broad attentions in recent years, which seeks for the approximate ideal solutions via decomposition of the spectral Laplacian. Nevertheless, it always suffers from the complexities of decomposition calculation, if large data is absorbed into clustering. 

On the other hand, some existing works consider the data analysis with missing values \cite{Rubin19SAMD}\cite{Dixon79PRPM}, which are in loss quite common during information convey and processing. The keypoints of such problem, have been reduced to supplement or fix the broken data with certain repatching approaches in theory, and the resulting data can be adopted to further handling \cite{Cheng15CRH}. Since data blocks are normally represented as numeric matrices, it is referred as a matrix completion problem, and solved with optimization tools \cite{Candes10MCN}\cite{Candes09EMC}. Nevertheless, the shortcoming of high complexities prohibits it from calculation efficiency, while the concrete framework is hardly to be further developed. More frequently, incomplete information often occurs in big data handling \cite{Nagashima20DI}\cite{Song20DI}, e.g., database and recommendation systems, another category of solutions are referred to fill the incomplete information with data imputation methods. Furthermore, data imputation is adopted to computer vision to enhance component analysis with missing values \cite{Chen04RMC}\cite{Shum95PCAMD}, and incremental learning of uncertain data has also been discussed \cite{Brand02ISVDMV}. As a consequence, some professional tools have been devised for data imputation, e.g., MICE \cite{Buuren11MICE}, Amelia II \cite{Honaker11Amelia}. In addition, multi-view clustering has attracted broad attentions in recent years \cite{Zhu19MVC}\cite{Lin21MVC}, which aims to learn the ideal partitions based on multiple representation of the common instances. It is noticeable that, the proposed method is also feasible for multi-view incomplete data analysis, owing to benefits of imputation models.

As a common limitation, clustering analysis of missing data are hardly to be addressed efficiently, as the two issues are independent problems and solved separately. Till now, there are quite a few solutions available for such issue, and most of them handle the clustering via a pre-step of data imputation \cite{Ma19OSI}\cite{Charles18SSC}. In order to improve the performance of data imputation approaches, there is a thirst for specific clustering solution to partitions of incomplete information \cite{Hruschka04EI}\cite{Boluki18OCMV}. Based on fuzzy \emph{c}-means clustering \cite{Cheng10MDP}, Hathaway and Bezdek proposed a generalized clustering framework for incomplete data \cite{Hathaway01FCI}. By devising a tolerant subspace clustering method, it is able to handle data with missing values flexibly \cite{Gunnemann11FFTSC}. In the literature, it has been reduced to be explained as optimization problems \cite{Zhang15MVSC}\cite{Kang17TL} with specific objective functions that are defined accordingly \cite{Goodfellow14GAN}.

In this work, a novel approach to clustering of incomplete data, termed leachable component clustering (LCC), is proposed. To distinguish from existing methods, the main contributions of this work are highlighted as below.
\begin{itemize}
\item The existing methods focus on representative clustering, while statistical patterns of incomplete data have been ignored. The proposed method seeks for optimized imputation, by exploiting the imputation models with respect to preservation of intrinsic distributions.
\item The proposed imputation models aim to fulfill the lost elements with the defined objectives, while fixed solutions can be obtained with calculation of the equations. Benefit from the efficiency of optimization, the ideal approximation can be calculated stably, while convergences can be achieved.
\item The proposed framework can be considered as a unified solution to analysis of incomplete data, and leads to the possible extensions of further advances.
\end{itemize}

The rest of this paper is organized as follows. The background knowledge of self-expressive clustering method is introduced in section II, and the basic conception of latent component models are given in section III. Then, the main idea of the proposed LCC is given in section IV. The experimental results on several incomplete data sets are given in section V. Finally, the conclusion is draw in section VI.

\section{Self-Expressive Clustering}
Given a set of data instances $ X = \left[ {{x_1},{x_2}, \cdots ,{x_n}} \right] \in {\mathds{R}^{d \times n}} $, and the amount of clustering $ c > 0 $. The standard \emph{k}-means clustering method attempts to assign each instance $ \begin{array}{*{20}{c}}
{{x_i},}&{i = 1, 2, \cdots , n}
\end{array} $ into $ c $ separate partitions, while the reconstruction errors of data can be minimized corresponding to the learned data means. 
Actually, it has been demonstrated that, it is equivalent for \emph{k}-means and line reconstruction, and can be summarized as a common objective with a perspective of quadratic optimization problem \cite{Bishop11PRML}. Afterward, such idea is summarized as a self-expressive property \cite{Zhang15MVSC}\cite{Kang17TL}, while appended extensions can be developed conveniently. Without loss of generality, the self-expressive problem can be defined as
\begin{equation}
  \begin{array}{ll}
{J_{SEC}} &  = \arg \mathop {\min }\limits_Z \frac{1}{2}{\left\| {X - XZ} \right\|^2} + \gamma \left\| Z \right\|_F^2	\\
{s.t.} & {{Z^T}1 = 1,}~~~{0 \le Z \le 1.}
\end{array}
\end{equation}
Here, $ Z \in \mathds{R}^{n \times n} $ indicates the coefficient matrix that is able to approximately reconstruct the original data instances, $ \gamma $ denotes the balance parameter, while the sub-conditions can avoid the trivial solution and ensure the positive of $ Z $. By extending the objective, the \emph{k}-means clustering is to optimize the following function, 
\begin{equation}
 {J_k} = \arg \mathop {\min }\limits_Z \frac{1}{2}{\left\| {X - {M_x}Z} \right\|^2} + \gamma \left\| Z \right\|_F^2,
\end{equation}
where $ M_x \in \mathds{R}^{d \times c} $ denotes the matrix consisting of data means of each cluster as one of its columns, and $ Z \in \mathds{R}^{c \times n} $ is the assigned indicator of partitions. Obviously, the \emph{k}-means clustering aims to optimize the objective function by seeking for the ideal $ Z $ and updating $ M_x $ iteratively, which can approximate the original $ X $.

\section{Latent Component Models of Data Imputation}
Data imputation aims to fullfil the missing values of data instances $ X $, and certain mechnism is adopted to approximately learn the predicated values that can approach to the ground truth $ \widetilde X $. 

\subsection{Principle Component Imputation Model}
The well-known principal component analysis (PCA) \cite{Turk91PCA} has been widely applied to learn the orthogonal components of the centered data block, and the main idea can be also explained as the ideal approximation of the original data in theory \cite{Cheng11MMC}\cite{Cheng18MMC}\cite{Golub13MC}. According to the self-expressive property, the orthogonal components of $ X $ and $ \widetilde X $ nearly share the common bases. Furthermore, the learned orthogonal components are able to make ideal reconstruction of $ X $, e.g.,
\begin{equation}
\begin{array}{ll}
{J_P}  &  = \arg \mathop {\min }\limits_P {\left\| {X - P{P^T}X} \right\|^2}		\\
{s.t.}&{{P^T}P = I},
\end{array}
\end{equation}
where $ P \in \mathds{R}^{d \times r} $ denotes the orthogonal bases of $ X $. As a consequence, the reconstructed data can be adopted to learn the following bases accordingly, and the stable approximation is able to be obtained. With the reconstructed patterns $ P{P^T}X $, it is believed that it holds the similar components with ground truth $ \widetilde X $, and the ideal imputation of missing values can be estimated.

\subsection{Self-Expressive Representation Learning}
Though incomplete data brings difficulties in information analysis, it is still possible to be analyzed while the dilemma of handling of original data can be avoided. 
Without loss of generality, representation learning attempts to learn the representative patterns of the original data with respect to specific objectives \cite{Saeed19MDS}\cite{Roweis00LLE}, e.g.,
\begin{equation}
 f:\begin{array}{*{20}{c}}
{{x_i} \to {y_i},}&{i = 1,2, \cdots ,n.}
\end{array}
\end{equation}
As a consequence, the learned $ y_i $ are adopted to further handling, which are more optimistic for incomplete representation \cite{Donoho06CS}\cite{Candes06Sparsity}. 

More specifically, the similarity objectives of instances are normally referred \cite{Cheng20AMKM}, and thus, the objective of pseudo data can be defined as
\begin{equation}\label{Eq-2}
 {J_S} = \arg \mathop {\min }\limits_y \frac{1}{2}\sum\limits_{i = 1}^n {\sum\limits_{j = 1}^n {{{\left\| {Sim\left( {\widehat {{x_{i,j} }},\widehat {{x_{j,i} }}} \right) - Sim\left( {{y_{i} },{y_{j} } } \right)} \right\|}^2}} }.
\end{equation}
Here, $ Sim\left( {\cdot},{\cdot} \right) $ denotes the similarity function between two instances. The $ \widehat x_{i, j} $ and $ \widehat x_{j, i} $ indicate the labeled instances corresponding to the valid features between two instances, which are respectively defined as
\begin{equation}
 \widehat {{x_{i,j} } } = {x_i} \odot {l_i} \odot {l_j}
\end{equation}
and
\begin{equation}
 \widehat {{x_{j,i} } } = {x_j} \odot {l_j} \odot {l_i},
\end{equation}
where $ \odot $ denotes the Hadamard product \cite{Golub13MC}.
Furthermore, the $ l_i $ denotes the labels that represent the validation and missing values of instance $ x_i $, e.g.,
\begin{equation}
 \begin{array}{l}
{l_{iq}} = \left\{ {\begin{array}{*{20}{c}}
{1,}&{if~{x_{iq}} \ne 0}\\
{0,}&{otherwise }.
\end{array}} \right.\\
q = 1, 2, \cdots , d.
\end{array}
\end{equation}
Here, $ x_{iq} $ denotes the $ q $-th element of the incomplete vector $ x_i $.
As a consequence, the learned $ y_i $ can approximately preserve the original similarities of data pairs. 

\subsection{Information Volume Preservation Model}
The main idea of information volume preservation (IVP) is based on the assumption that, the informative patterns of data actually hold a nearly constant volume corresponding to different partitions. Contrarily, IVP is able to contribute to pattern unfolding and distribution learning \cite{Cheng20IVP}\cite{Cheng21FA}. In terms of this idea, it is to reserve the information contained in ground truth to be approximate to incomplete data, while reconstruction is adopted to estimate $ {\widetilde {x_i}} $. 

With respect to IVP model, the feature distances of incomplete data can be estimated and nearly approximate to the characteristics of ground truth, as the lost patterns can be weakened in the high-dimensional feature space. Particularly, the reconstruction of kernel components are adopted, and the information volumes $ {V_{f}}\left( {{x_i}} \right) $ of kernels $ h \left( \cdot, \cdot \right) $ associated with incomplete data $ x_i,~~ i = 1,2, \cdots, n $, can be defined as
\begin{equation}
\begin{array}{ll}
   &  {V_{f}}\left( {{x_i}} \right)   = \sum\limits_{j = 1}^k {h\left( {{x_i \odot {l_i} \odot {l_j}}, {x_j \odot {l_i} \odot {l_j}}} \right)} ,		\\
   &  i  = 1, 2, \cdots , n.
\end{array}
\end{equation}
Here, $ x_j $ denotes the $ k $ nearest neighbors of $ x_i $.
As a consequence, the dilemma of incomplete patterns can be alleviated, and IVP aims to preserve the obtained volumes in the fulfilled data that are to be estimated. Furthermore, it is necessary to make the data patterns be the explicit characteristics of normal instances, which can be further depicted with data similarities, e.g., euclidean or cosine distances. Accordingly,  the objective of IVP model can be defined as
\begin{equation}
  \begin{array}{*{20}{c}}
{{J_{IVP} } = \arg \mathop {\min }\limits_{\widetilde {{x_i}}} \left| {V\left( { \widetilde{x_i} } \right) - {V_{f}}\left( {{x_i}} \right)} \right|,}&{i = 1,2, \cdots , n.}
\end{array}
\end{equation}
Thus, the ideal approximation can be learned, by solving the equations and repairing incomplete data alternately.

\section{Leachable Component Learning}
Derived from information preservation model, a novel approach to data imputation is devised in this work. And the estimation of data distributions are exploited to predicate the lost patterns of incomplete data.

\subsection{Bayes Alignment Model}
Assume that the distribution parameters of fulfilled data are equivalent to the ones of ground truth, then it is feasible to learn the imputation by solving the latent equations of associated probabilistic models \cite{Hastie11ESL}. More specifically, the basic idea of the imputation model is based on the assumption that, the difference of data distributions holds equivalence between the incomplete and fulfilled instances, which can approximate to the distribution of ground truth. Distinguishingly, the Bayes alignment in the context, actually stands for the affiliation probabilities of each instance associated with certain data partitions, and are approximately represented by distributions. Nevertheless, it is worthwhile to highlight several calculation issues firstly. The sample means are to be calculated associated with the available values of incomplete data, and formally, the valid values of each instance is referred, which is defined as
\begin{equation}
\begin{array}{ll}
{{m_i}  = \frac{1}{{{\delta _i}}}\sum\limits_{j = 1}^n {{x_{ji}} \cdot {l_{ji}}} ,} & {i = 1,2, \cdots ,d.}	\\
{\delta _i}  = \sum\limits_{j = 1}^n {{l_{ji}}} &
\end{array}
\end{equation}
Accordingly, the sample variance $ \sigma $ is to be calculated associated with the valid labels of data features. Then, the distribution of valid features of each data can be calculated according to normal distribution of samples. As a consequence, the distribution of samples with respect to a specific data $ x_i $ can be approximately calculated as
\begin{equation}
{G_{m,{\sigma ^2}}}\left( {{x_i}} \right) = \frac{1}{{\sqrt {2\pi } \sigma }}\exp \left( { - \frac{{{{\left( {{x_i} - m \odot {l_i} } \right)}^T}\left( {{x_i} - m  \odot {l_i}} \right)}}{{2{\sigma ^2}}}} \right).
\end{equation}
Note that, the obtained distribution $ {G_{m,{\sigma ^2}}}\left( {{x_i}} \right) $ is an estimated value of incomplete data associated with the available characteristics of data patterns.

As a consequence, it attempts to learn the ideal approximation $ \widetilde{x_i} $ that can reserve the statistical distributions of each incomplete data, such as
\begin{equation}
 \begin{array}{*{20}{c}}
{{J_{BA} } = \arg \mathop {\min }\limits_{\widetilde {{x_i}}} \left| {{G_{\widetilde m,\widetilde {{\sigma ^2}}}}\left( {\widetilde {{x_i}}} \right) - {G_{m,{\sigma ^2}}}\left( {{x_i}} \right)} \right|,}&{i = 1,2, \cdots ,n.}
\end{array}
\end{equation}
Furthermore, the statistical parameters of fulfilled data can be calculated naturally, and obtain the approximate $ \widetilde{m} $ and $ \widetilde{\sigma^2} $. There are several available calculation solutions to obtain the latent values of incomplete data with respect to the objective of $ J_{BA} $. Nevertheless, the objective of Bayes alignment imputation model can be solved with the simple calculation of equations. 


\subsection{Solution of Latent Leachable Learning}
The leachable imputation models aim to learn the approximate incomplete data $ x_i, ~ i = 1, 2, \cdots, n $ with objectives of latent components. 
For the Bayes alignment model, it is to solve the following equation for each incomplete data,
\begin{equation}\label{Eq-1}
 \widetilde {{x_i}} \cdot \widetilde {{x_i}} - 2\widetilde m \cdot \widetilde {{x_i}} + \widetilde m \cdot \widetilde m + 2{\widetilde \sigma ^2}\log \left( {\sqrt {2\pi } \widetilde \sigma {G_{m,{\sigma ^2}}}\left( {{x_i}} \right)} \right) = 0.
\end{equation}
Note that, the objective is a standard quadratic equation, but the solution to the above equation normally exists in complex domain. 
Nevertheless, it can be approximated in real values with numeric rotations, and achievements are promised to be obtained with efficiency. 

As a consequence, the ideal solution $ \widetilde{x_i} $ to the proposed imputation model can be calculated as
\begin{equation}\label{Eq-4}
\widetilde {{x_i}} = \widetilde m \pm \sqrt {2{{\widetilde \sigma }^2}\tau \left( {\log \left( {\sqrt {2\pi } \widetilde \sigma {G_{m,{\sigma ^2}}}\left( {{x_i}} \right)} \right)} \right)}.
\end{equation}
Here, $ \tau \left( \cdot \right) $ denotes the absolute function, such as
\begin{equation}
\tau \left( x \right) = \left\{ {\begin{array}{*{20}{c}}
{\left| x \right|,}&{if~x \ne 0}\\
{0,}&{otherwise}.
\end{array}} \right.
\end{equation} 
In addition, there obtains two alternative $ \widetilde{x_i} $ for approximation in either model. During the first iteration, $ \widetilde{x_i} $ that is close to mean $ \widetilde{m} $ is chosen as the solution of equation. Then, the ideal $ \widetilde{x_i} $ is to be updated as the one that is close to the obtained $ \widetilde{x_i} $ in the last iteration. Nevertheless, it has been shown that either $ \widetilde{x_i} $ can be competent to learn the ideal results. After obtaining the approximate data, it is to update the parameters and repeatedly optimize the latent components during iterations.

Furthermore, the proposed method is able to be recycled, due to its intrinsic relationship with data means. More specifically, the learned partitions can be the supposed affiliation of each instance as learning models in the next cycle, and the leachable components can be achieved with the fresh data groups. In other words, the parameters of distributions can be estimated and updated with the learned groups during each iteration, with respect to the instances that belong to the common partitions, as well as the lost patterns of instances. As a consequence, the fulfilled patterns can be achieved with repeated cycles, benefiting from the initially leachable learning.

\subsection{Discussion}
Compared with existing methods, the fixed solutions can be obtained with the proposed alignment model during each iteration, and the improved results are available in the further optimization steps. Nevertheless, the optimization convergence is hardly to be achieved, while the supplemented data can be approximated to the original data with respect to global distribution. As a consequence, the similarity function can be defined in accordance with Eq. (\ref{Eq-1}), and the lost elements can be estimated by solving the approximate equation. Note that, the proposed supplementation models mainly rely on the solutions to line equations, which can be efficiently calculated with low computational complexities.

Similarly, the original IVP imputation model aims to solve the equation associated with the reservation of information volumes, e.g.,
\begin{equation}\label{Eq-2}
 \begin{array}{l}
\sum\limits_{j = 1}^k {\left( {\widetilde {{x_i}} \cdot \widetilde {{x_i}} - 2 \cdot {x_j} \cdot \widetilde {{x_i}} + {x_j} \cdot {x_j}} \right)}  - \sum\limits_{j = 1}^k {{S_f}\left( {{x_i},{x_j}} \right)} \\
 = k \cdot \widetilde {{x_i}} \cdot \widetilde {{x_i}} - 2 \cdot \widetilde {x{}_i} \cdot \sum\limits_{j = 1}^k {{x_j}}  + \sum\limits_{j = 1}^k {{x_j} \cdot {x_j}}  - \sum\limits_{j = 1}^k {{S_f}\left( {{x_i},{x_j}} \right)} \\
 = 0
\end{array}
\end{equation}
where $ S_{f} $ denotes the similarities of incomplete data between $ x_i $ and $ x_j $ in feature space, which is defined as
\begin{equation}
  {S_f}\left( {{x_i},{x_j}} \right) = Sim_f \left( {\widehat {{x_i}},\widehat {{x_j}}} \right).
\end{equation}
Nevertheless, it is hardly to achieve convergence in a common step, owing to independent optimization of each instance, which is necessary to calculate the information volumes associated with each instance. In practice, it is alleviated by setting an upper bound of iterations, and sampling is adopted to reduce the complexities.

%
%
%

\begin{figure*}
    \centering
    \subfigure[]{ \includegraphics[width=0.23\textwidth]{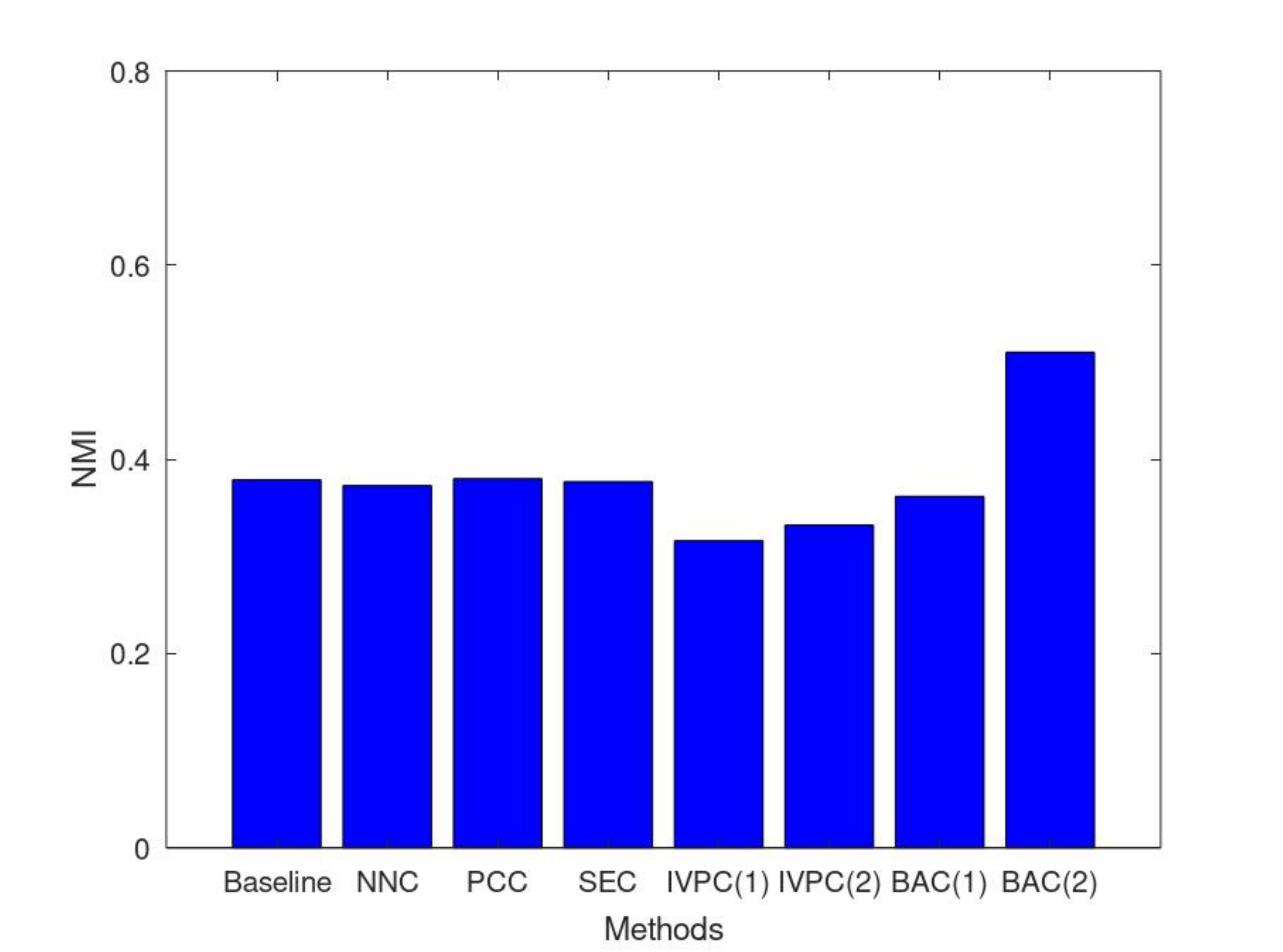}	}
    \subfigure[]{ \includegraphics[width=0.23\textwidth]{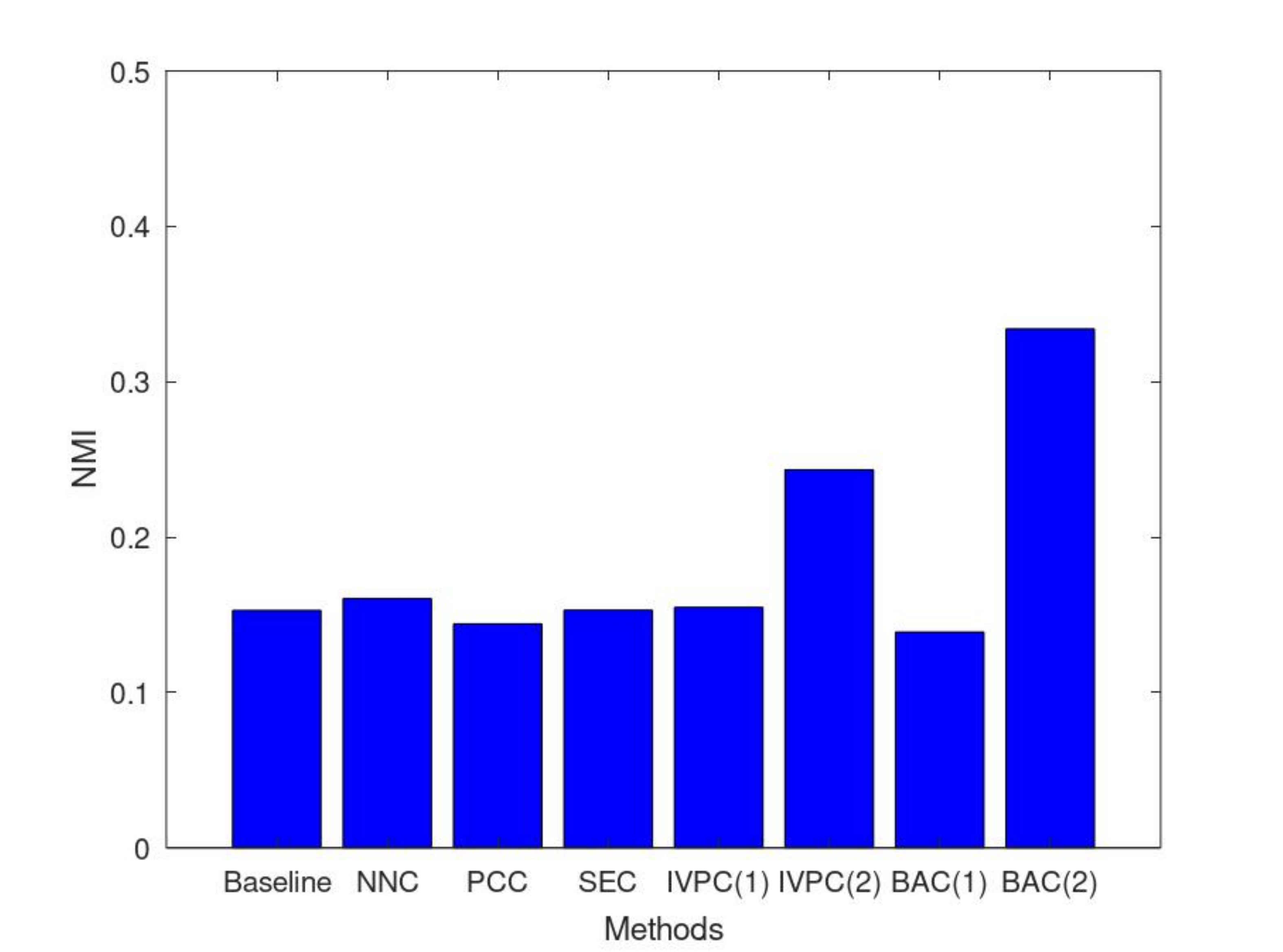}	}
    \subfigure[]{ \includegraphics[width=0.23\textwidth]{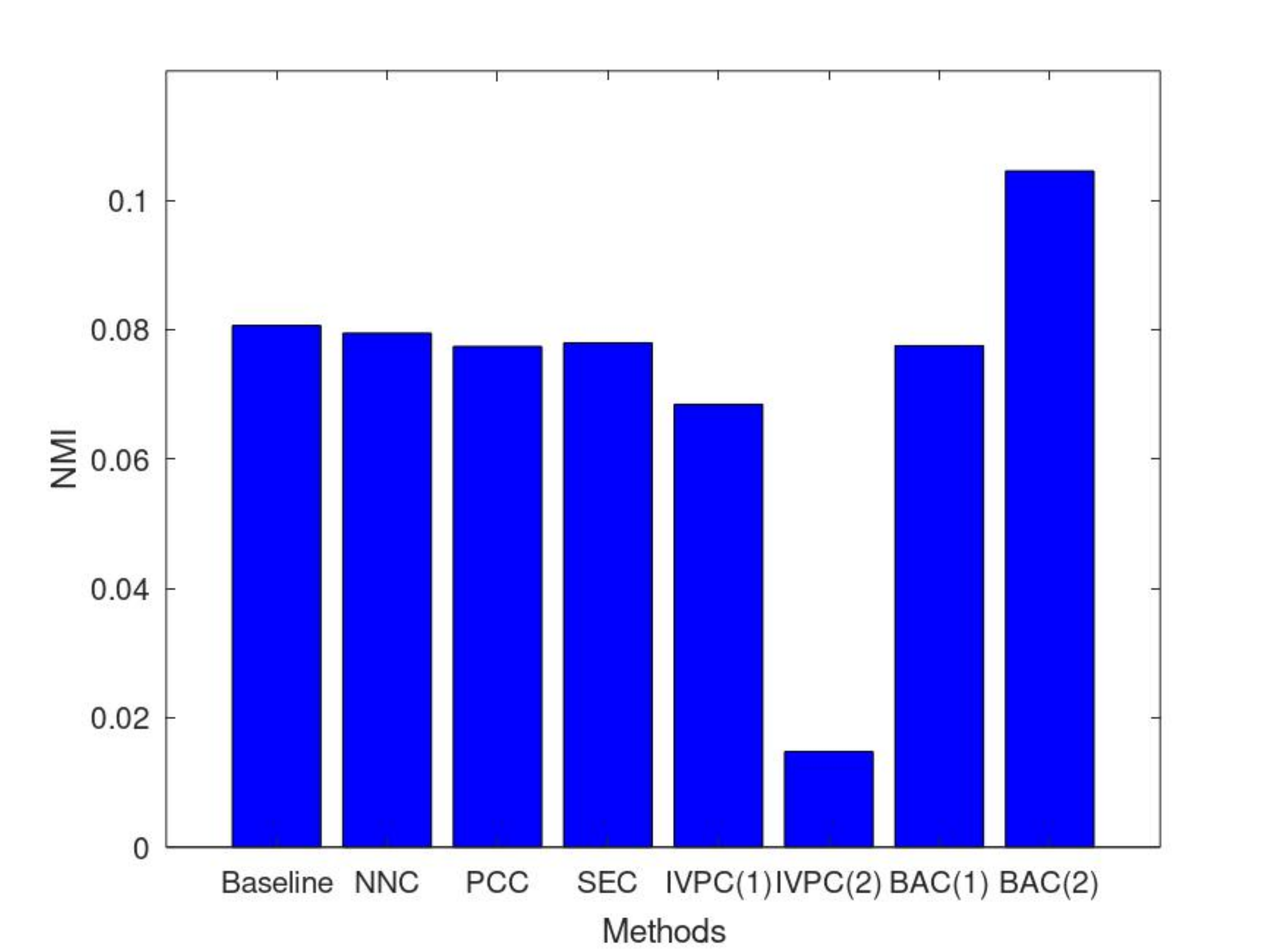}	}
    \subfigure[]{ \includegraphics[width=0.23\textwidth]{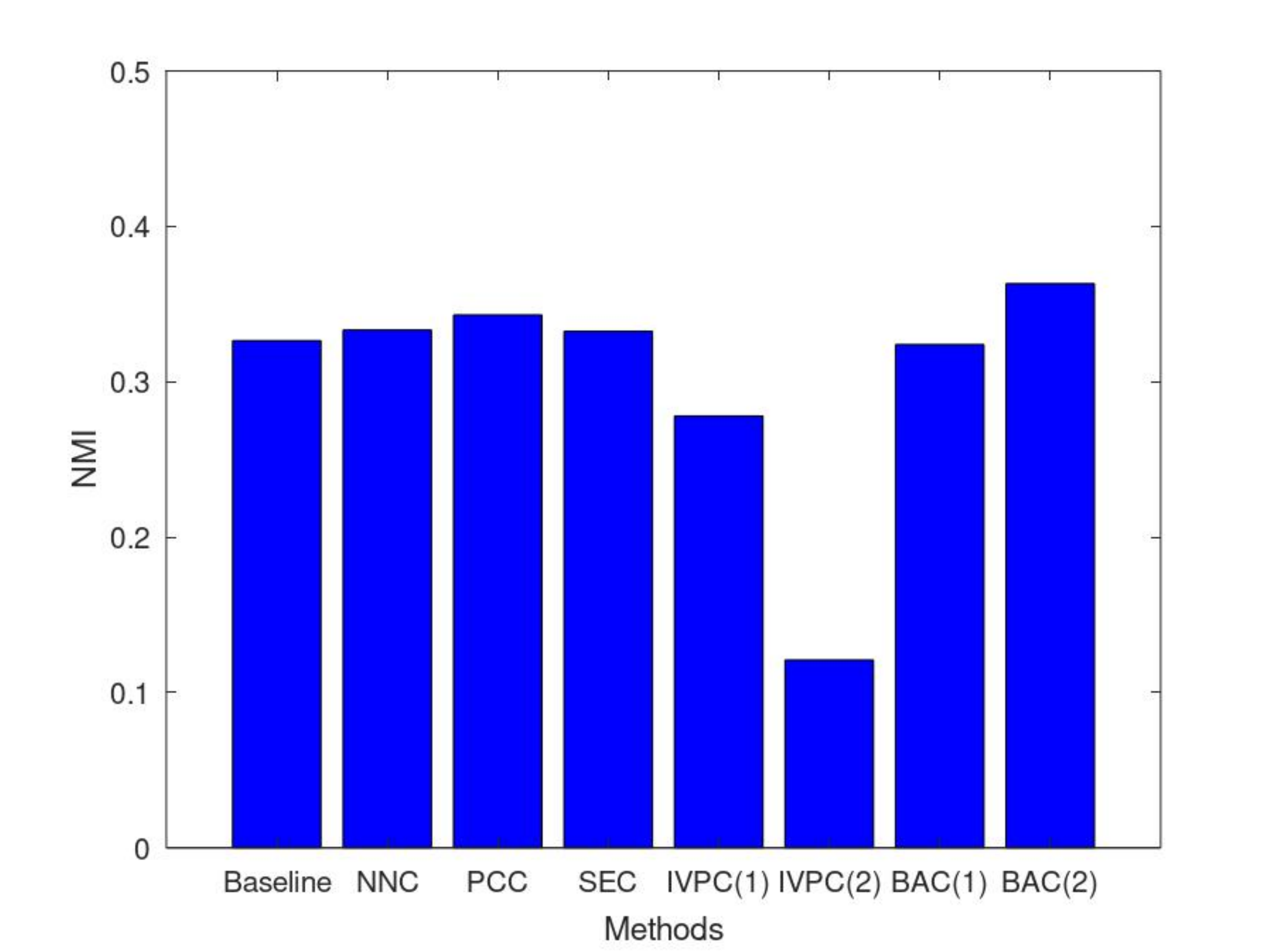}	}
    \caption{The obtained Normalized Mutual Information (NMI) of different algorithms on four data sets. (a) Birds (b) Firewall (c) Flower (d) Monkey.	}
\end{figure*}
\begin{figure*}
    \centering
    \subfigure[]{ \includegraphics[width=0.23\textwidth]{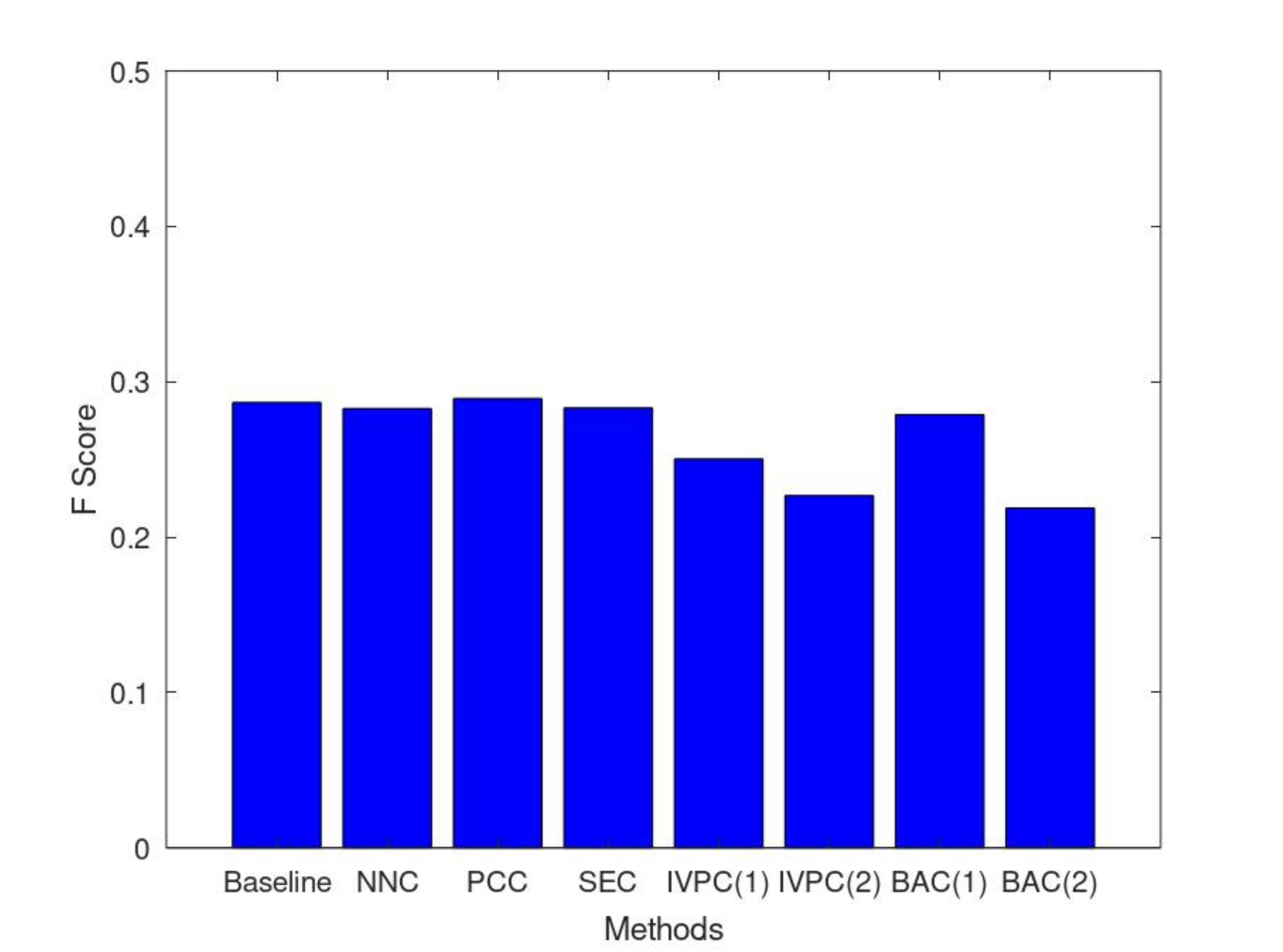}	}
    \subfigure[]{ \includegraphics[width=0.23\textwidth]{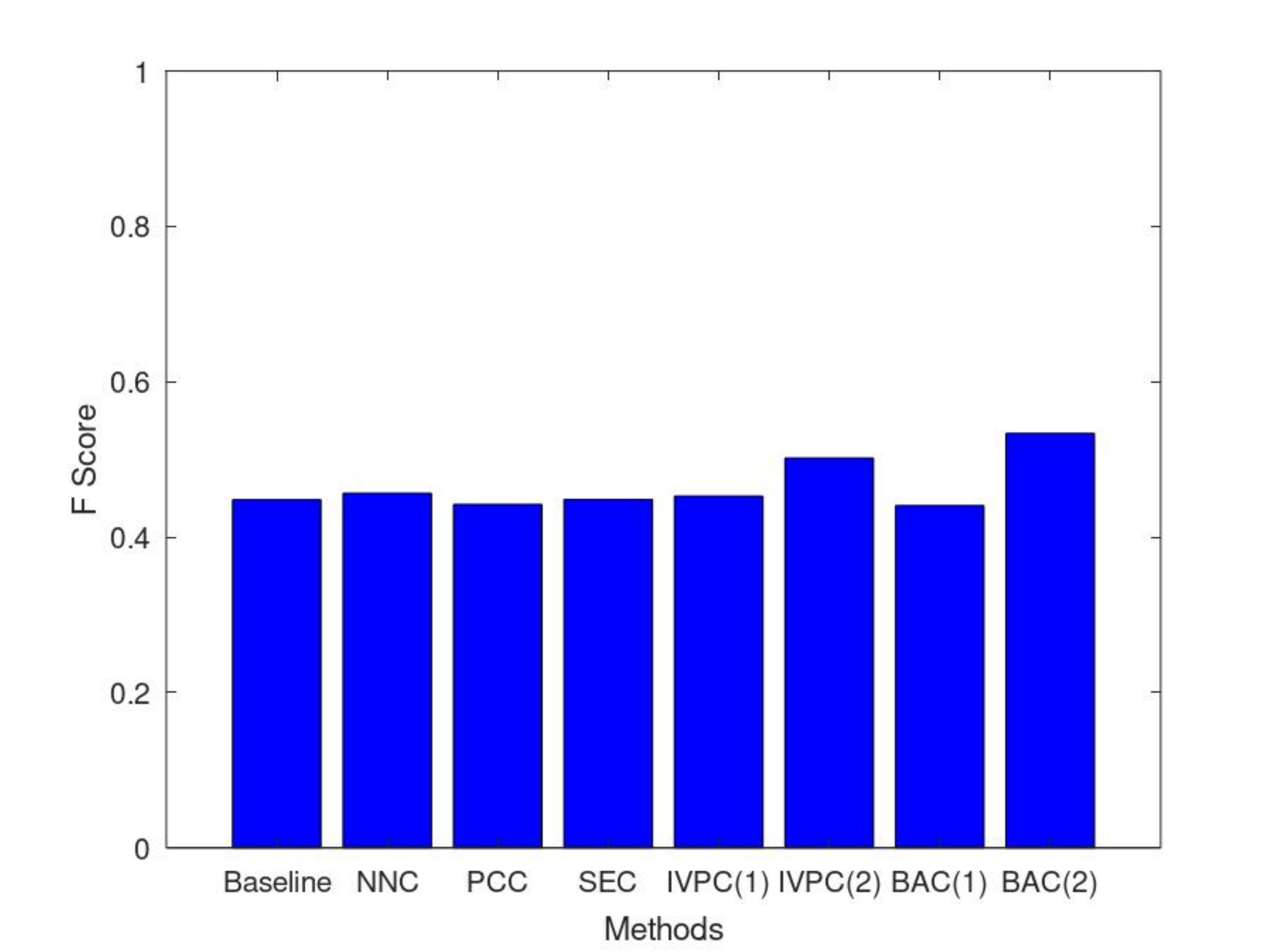}	}
    \subfigure[]{ \includegraphics[width=0.23\textwidth]{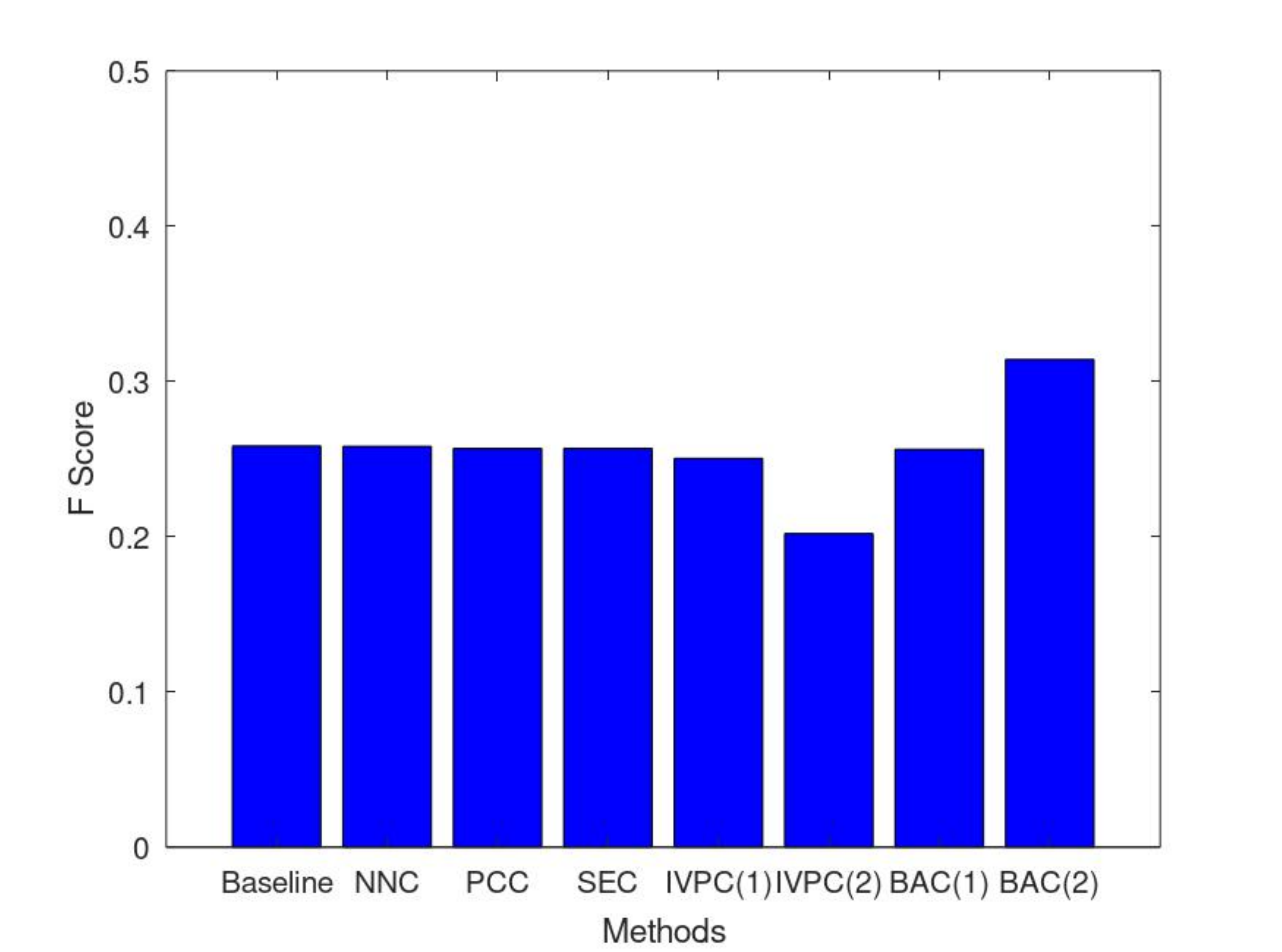}	}
    \subfigure[]{ \includegraphics[width=0.23\textwidth]{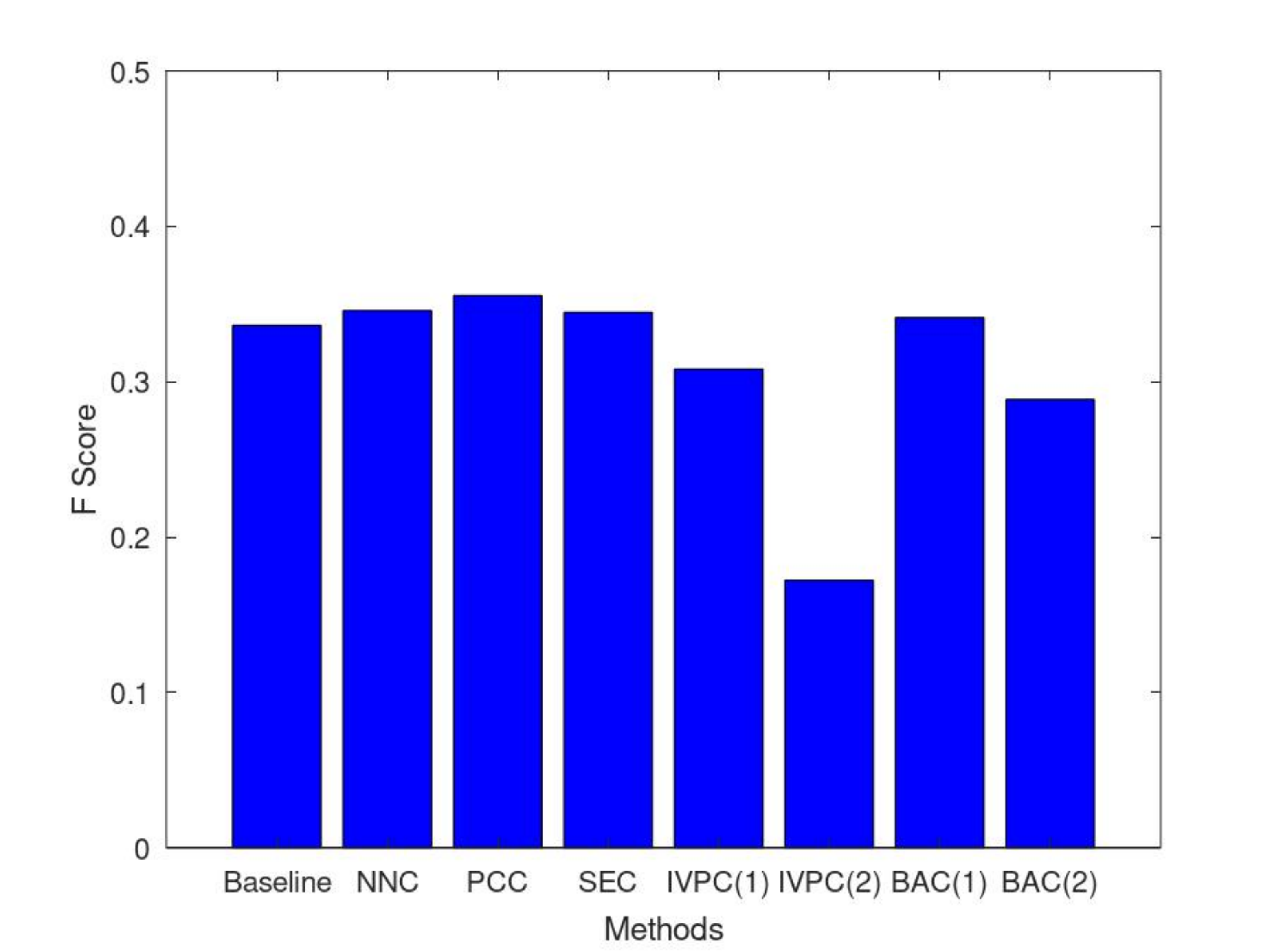}	}
    \caption{The obtained F Score of different algorithms on four data sets. (a) Birds (b) Firewall (c) Flower (d) Monkey.	 }
\end{figure*}
\begin{figure*}
    \centering
    \subfigure[]{ \includegraphics[width=0.23\textwidth]{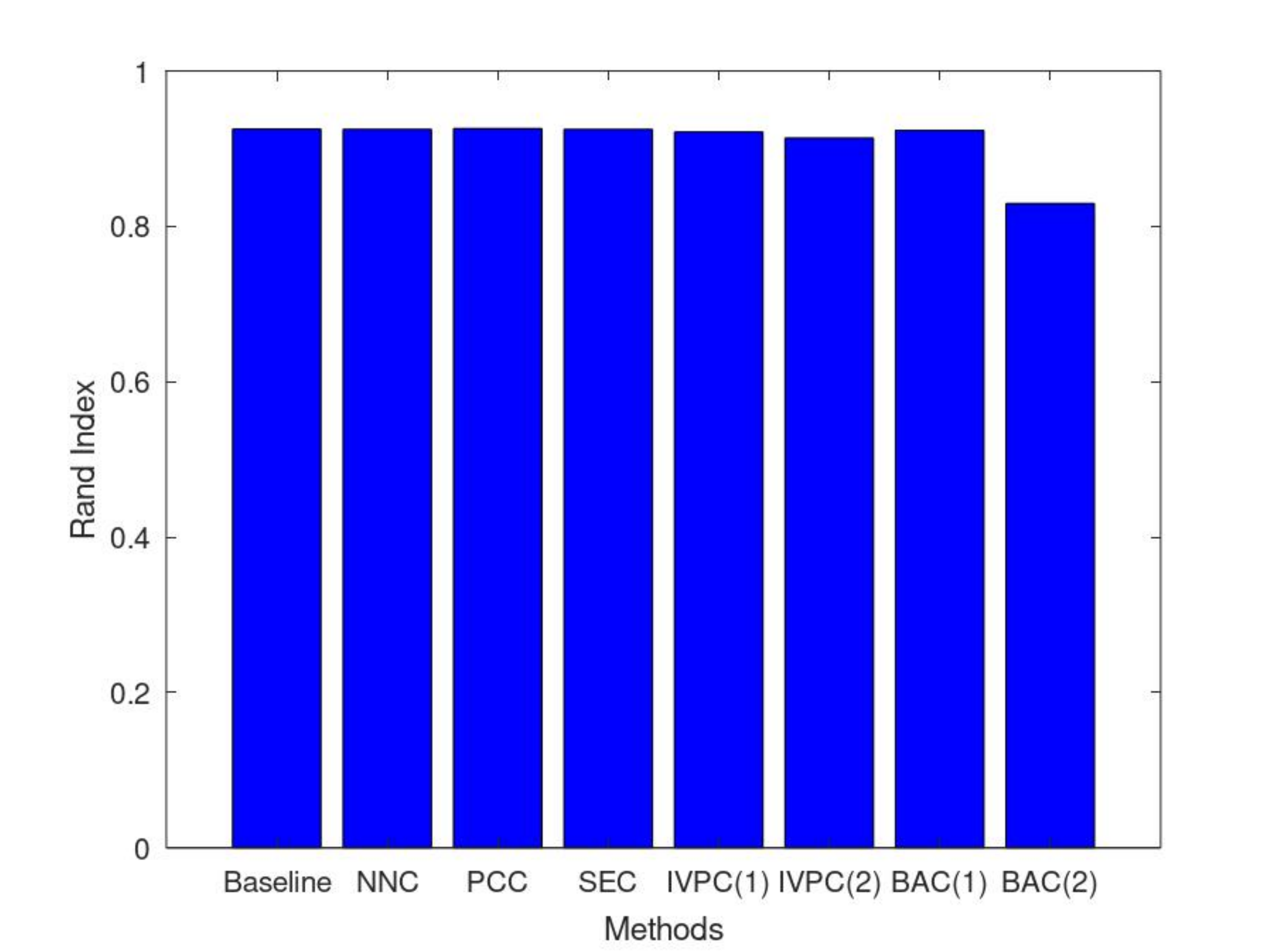}		}
    \subfigure[]{ \includegraphics[width=0.23\textwidth]{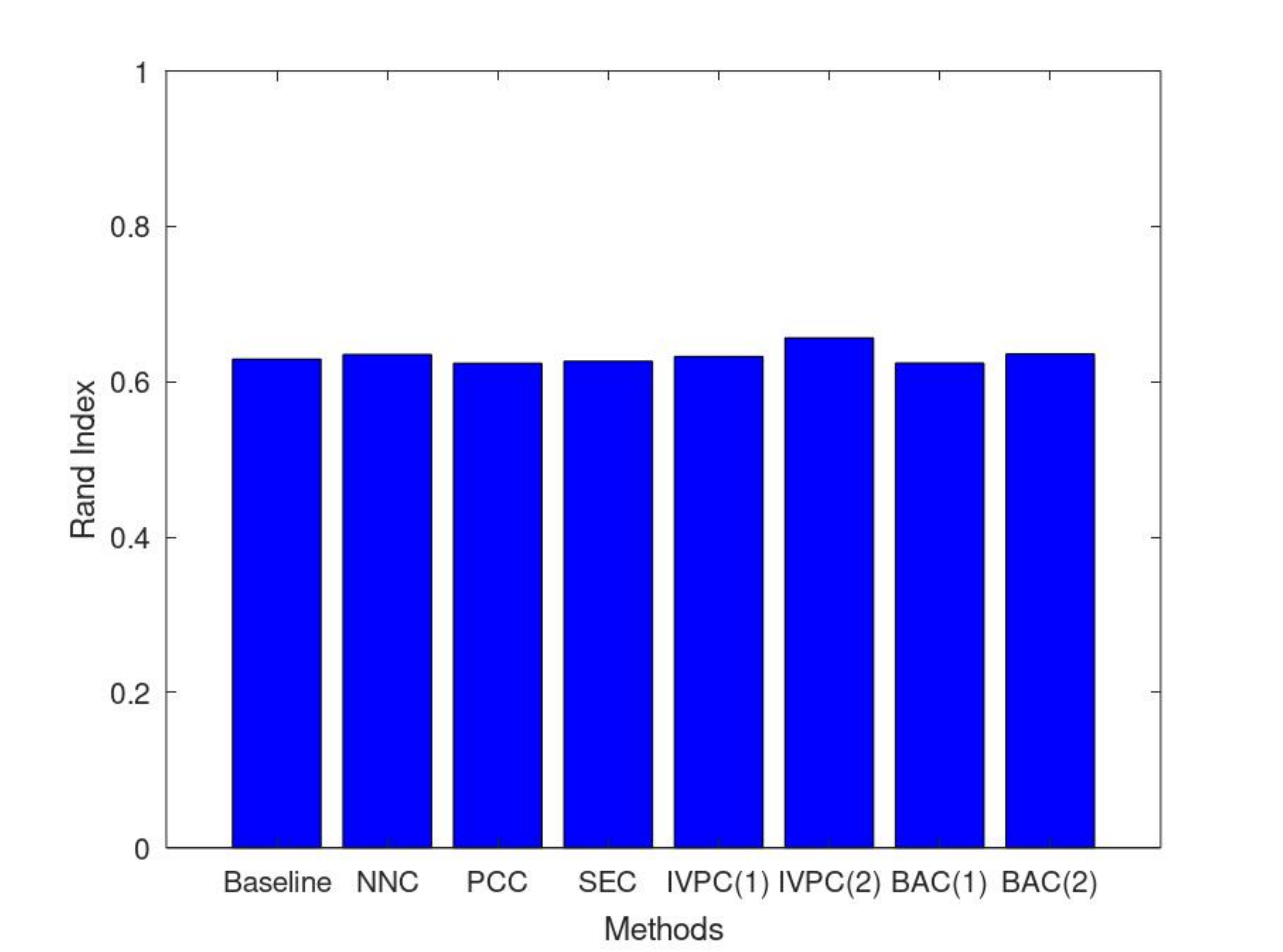}	}
    \subfigure[]{ \includegraphics[width=0.23\textwidth]{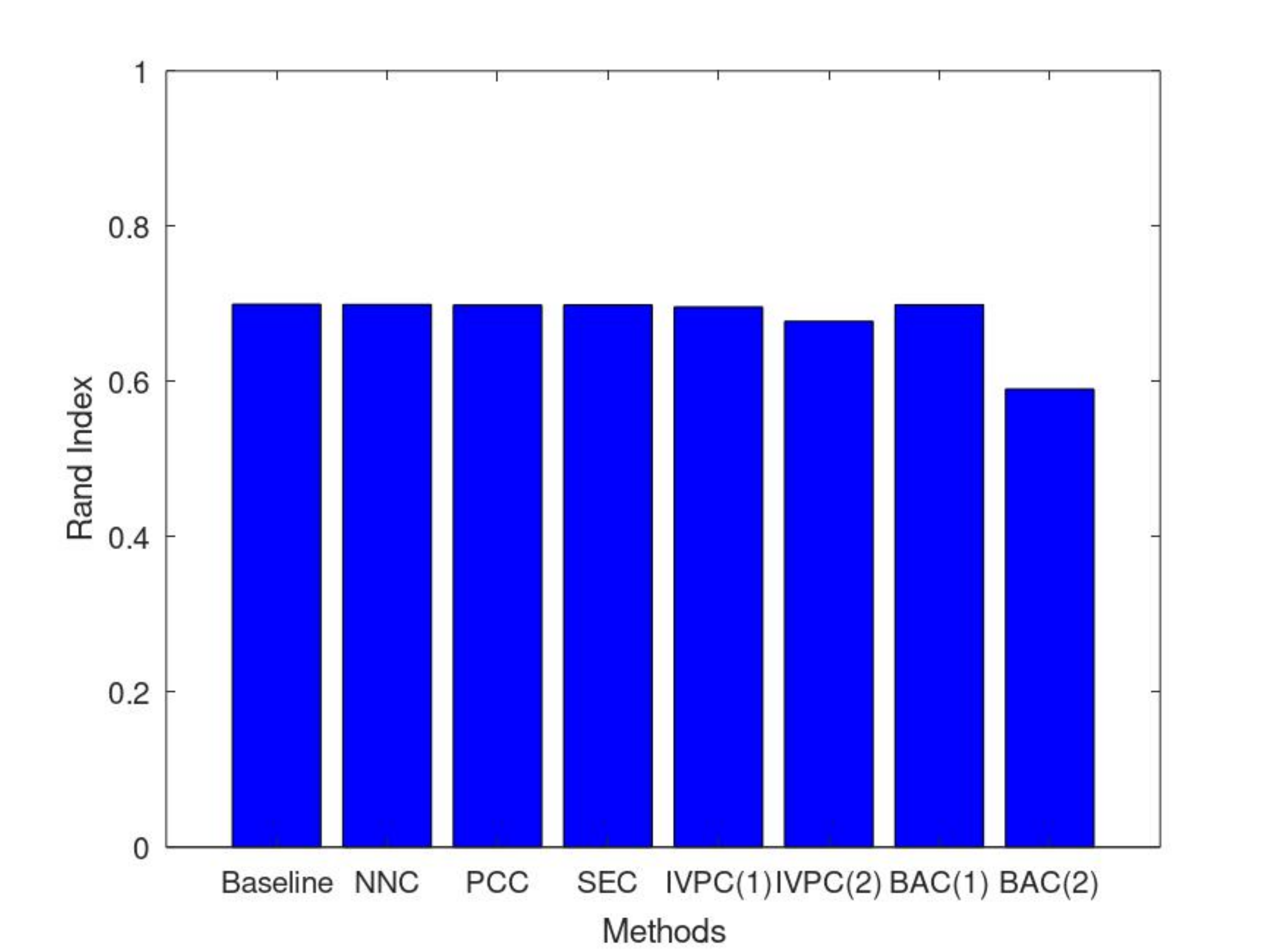}		}
    \subfigure[]{ \includegraphics[width=0.23\textwidth]{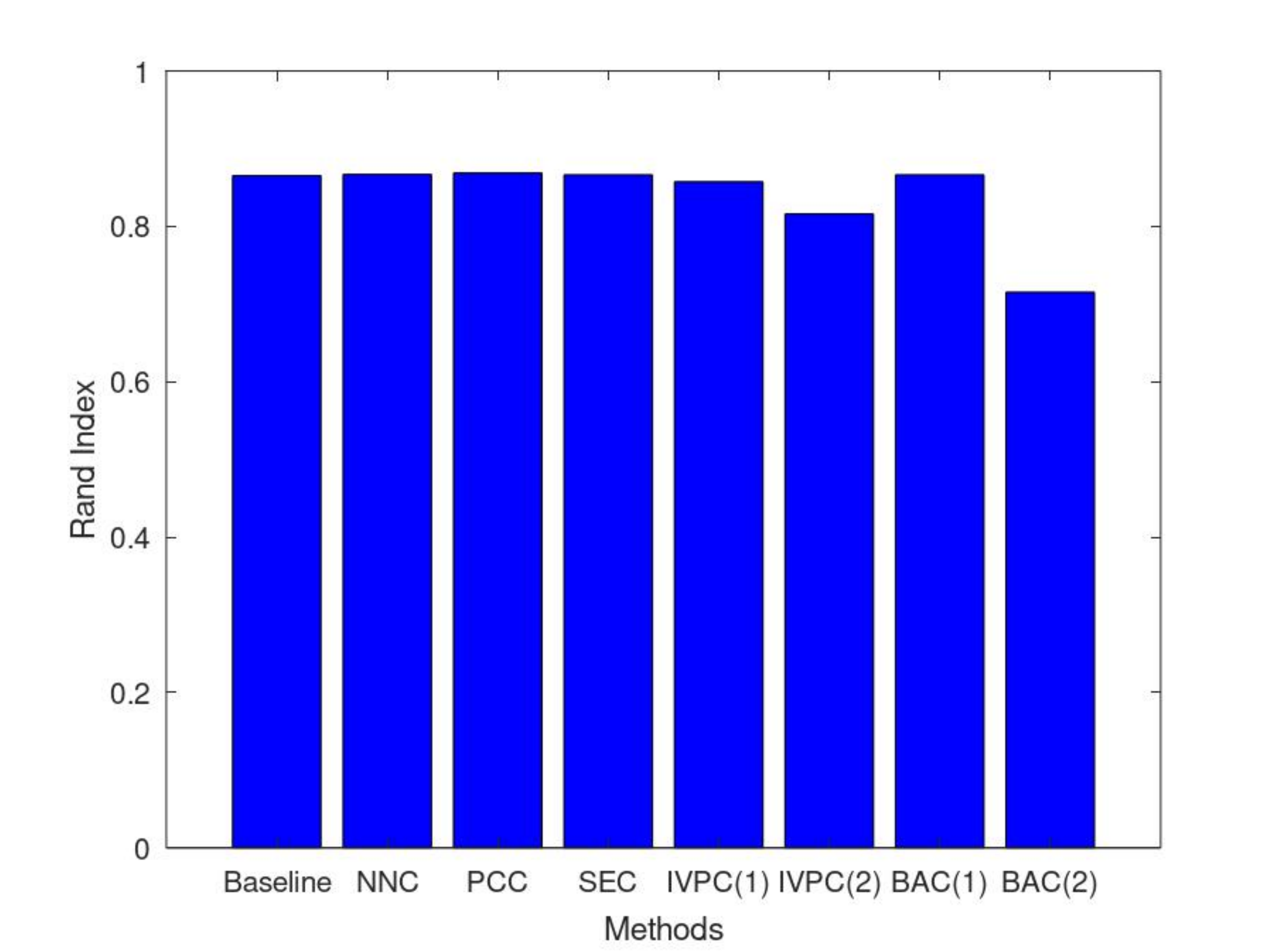}	}
    \caption{The obtained Rand Index of different algorithms on four data sets. (a) Birds (b) Firewall (c) Flower (d) Monkey.	 }
\end{figure*}
\begin{table}[!t]
\centering
\caption{The details of data sets} \centering
\begin{tabular}{c|c|c|c}\hline\hline
  & & &   	\\[-6pt]
Data & Partition & Dimensionality & Instance 		\\
  & & &   	\\[-8pt] \hline
  Birds & 20 & 50 & 3625 	\\
  Firewall & 3 & 11 & 3000  	\\
  Flower & 5 & 50 & 4323  	\\
  Monkey Species & 10 & 50 & 1098 	\\
\hline
\end{tabular}
\end{table}
\section{Experiments}
In this section, the performance of the proposed methods are to be evaluated on several artificial data sets. Four data sets are employed in the experiments, including Internet Firewall\footnote{http://archive.ics.uci.edu/ml/datasets/Internet+Firewall+Data}, 100 Birds\footnote{https://www.kaggle.com/gpiosenka/100-bird-species}, Flower Recognition\footnote{https://www.kaggle.com/alxmamaev/flowers-recognition}, 10 Monkey Species\footnote{https://www.kaggle.com/slothkong/10-monkey-species}. 
All experiments are performed on the hardware of 2.9 GHz CPU with six cores and 16 GB RAM. During the experiments, partial instances of some data sets are employed, and the deep and reduced representations of image data sets are extracted, which consist of normalized patterns of 50 dimensions for each instance. For the Internet Firewall data set, the 1,000 instances of each category are randomly selected, and the image data of 20 categories in bird species are selected. In summary, the details of involved data sets are given in Tab. I. 
Several state-of-the-art methods associated with data imputation of latent component models are involved to make a comparison of clustering performance, which are given as follows.
\begin{itemize}
\item Baseline \emph{k}-means clustering \cite{Kanungo02KMeans}: The \emph{k}-means clustering on raw incomplete data.
\item Nearest neighborhood fulfilled clustering  (NNC) \cite{Bishop11PRML}: The NN imputation based clustering of incomplete data. To reduce complexity, the average fulfillment from three random neighbors are preferred in the experiments.
\item Principle components fulfilled clustering (PCC) \cite{Cheng11MMC}\cite{Cheng18MMC}\cite{Shum95PCAMD}: The PC imputation based clustering of incomplete data.
\item Self-expressive clustering (SEC) with incomplete data \cite{Zhang15MVSC}\cite{Kang17TL}: The SEC method on raw incomplete data.
\item Information volume preservation fulfilled clustering, namely IVPC(1) and IVPC(2): The proposed IVP imputation followed by the standard self-expressive and \emph{k}-means clustering respectively.
\item Bayes alignment fulfilled clustering, namely BAC(1) and BAC(2): The BA imputation followed by the self-expressive and \emph{k}-means clustering respectively.
\end{itemize}

For each data set, \emph{thirty} percent of whole elements are randomly selected and set to be null to make the incomplete data. Then, the data supplementation and standard SEC are performed by following each algorithm. Note that, the initial partitions of instances are quite important for optimized clustering, while random initialization is employed in the experiments. To alleviate the occasional sense, all experiments are repeated \emph{five} times and the \emph{average} results are calculated and recorded as the outputs. The obtained results of the first experiment are given in Fig. 1-3, corresponding to three clustering measure.

According to the experimental results, the best NMI results are obtained by BAC method on all four data sets, while better outputs are achieved with F measure. The obtained results of NNC and PCC are quite similar to each other, as well as the Baseline. In other words, these two methods produce the similar fulfillment for data imputation, and close performance are given for clustering. Furthermore, the outputs of IVPC are different from each other, corresponding to different clustering mechanism. And totally, \emph{k}-means based IVPC is able to give better partitions compared with SEC in most cases, especially on the Firewall and Flower data sets. Though \emph{k}-means based BAC is better with NMI and F measures on Firewall and Flower, the inferior results are obtained if the Rand Index measure is referred. Nevertheless, it is noticeable that, the obtained Rand Index of different methods are quite close to each other, and it is hardly to make conclusion.
\begin{figure*}
    \centering
    \subfigure[]{ \includegraphics[width=0.23\textwidth]{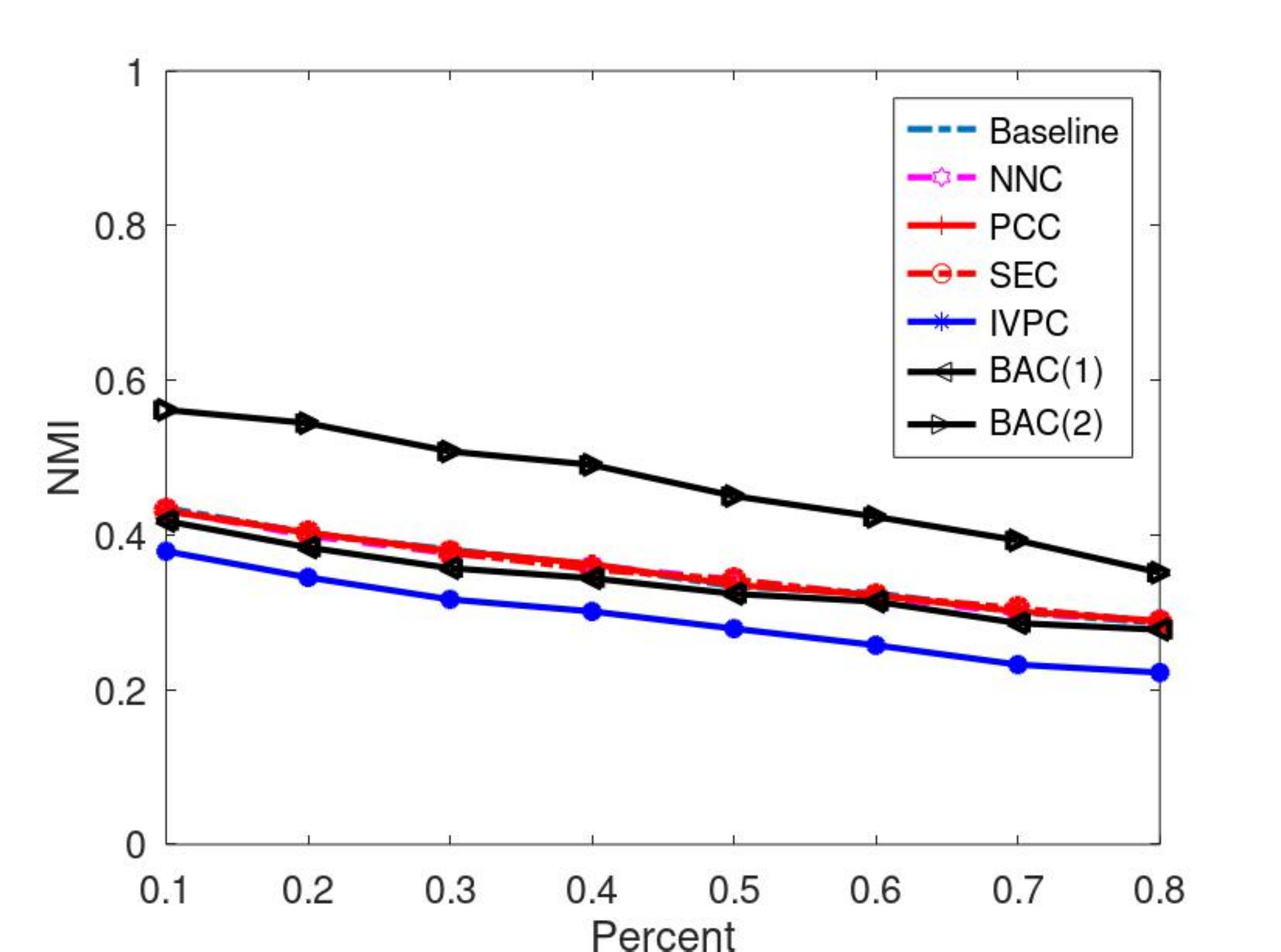}		}
    \subfigure[]{ \includegraphics[width=0.23\textwidth]{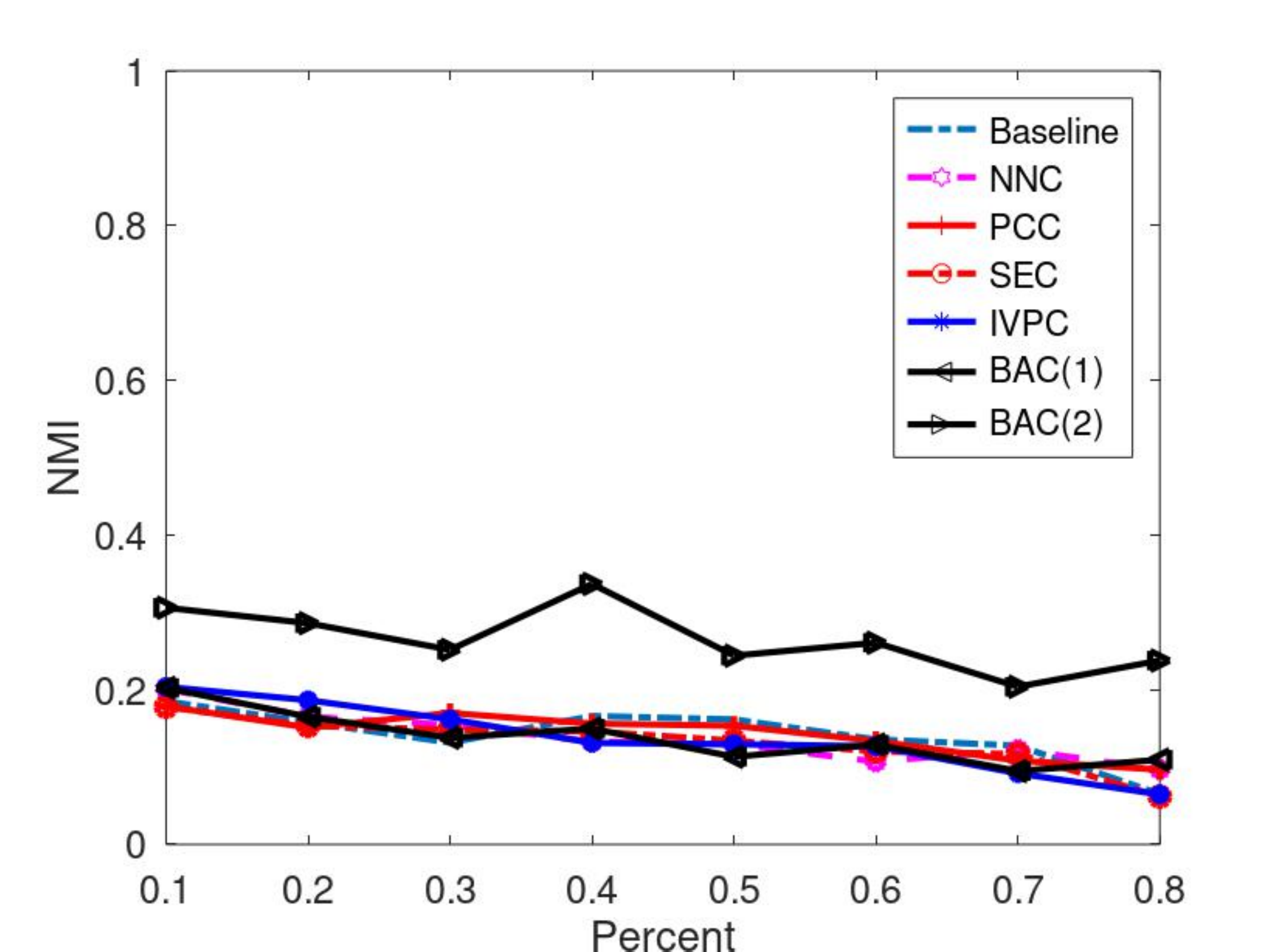}	}
    \subfigure[]{ \includegraphics[width=0.23\textwidth]{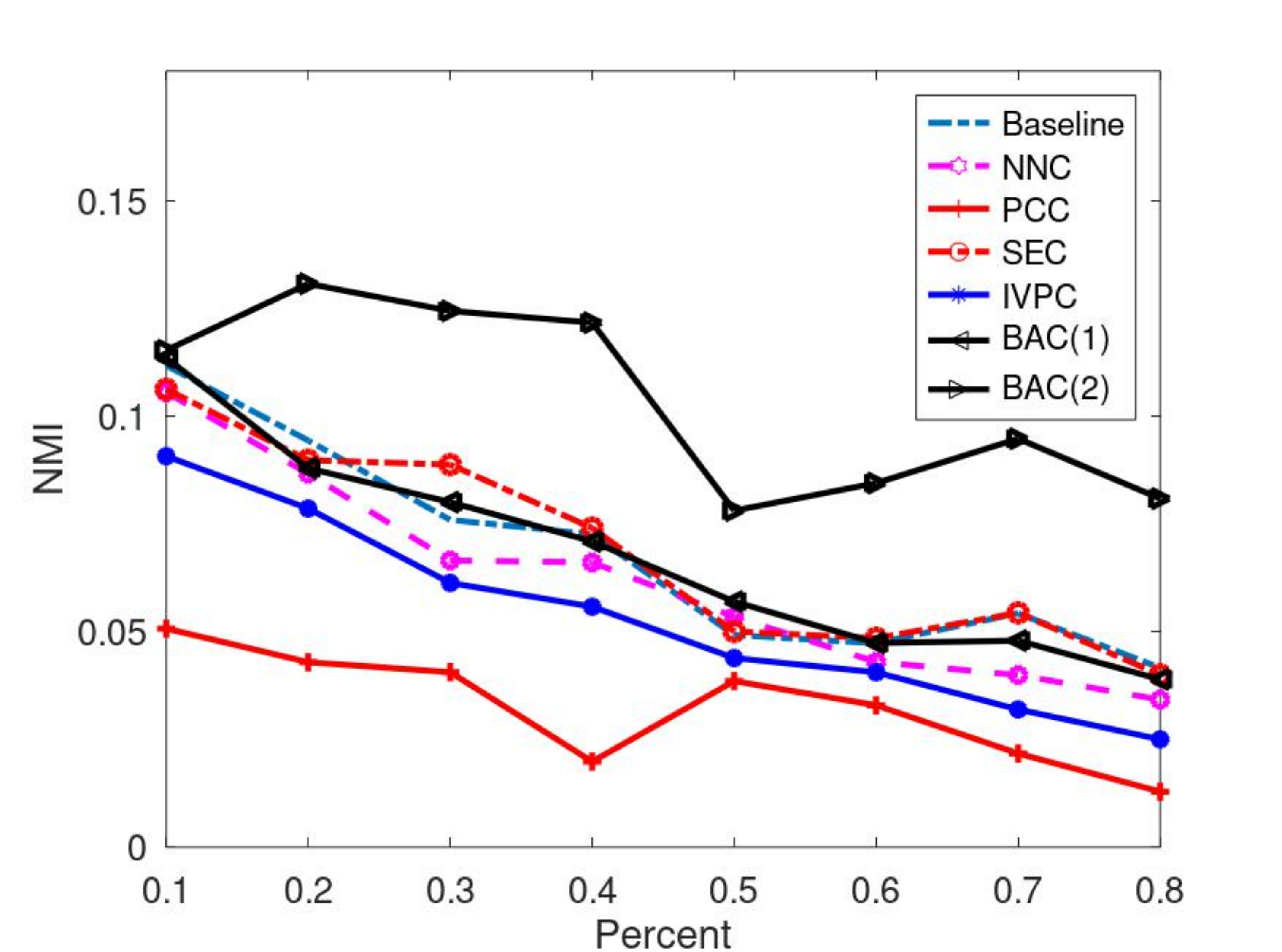}		}
    \subfigure[]{ \includegraphics[width=0.23\textwidth]{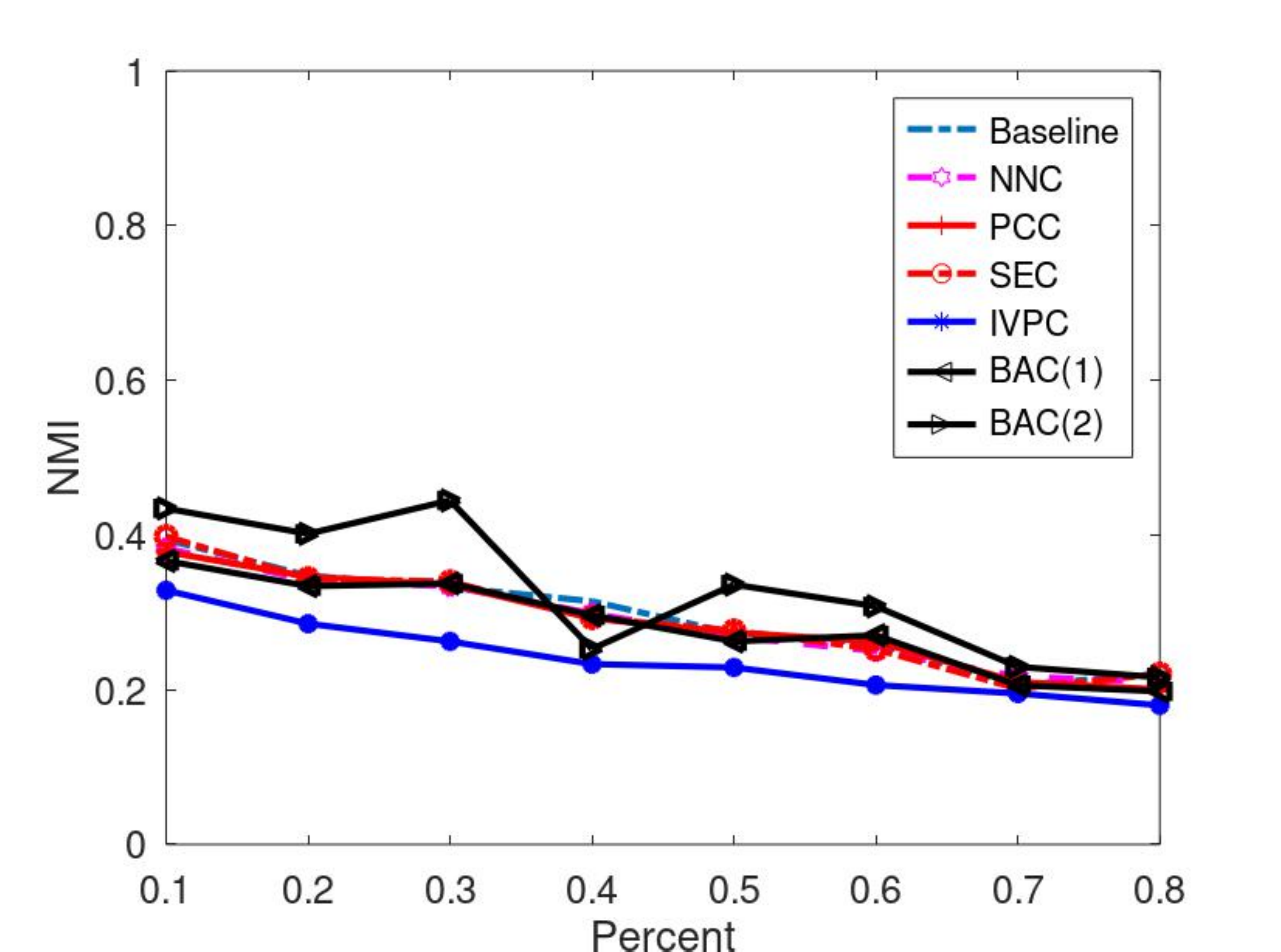}	}
    \caption{The obtained Normalized Mutual Information (NMI) of different algorithms associated with different percent of null elements on four data sets. (a) Birds (b) Firewall (c) Flower (d) Monkey.	 }
\end{figure*}
\begin{figure*}
    \centering
    \subfigure[]{ \includegraphics[width=0.23\textwidth]{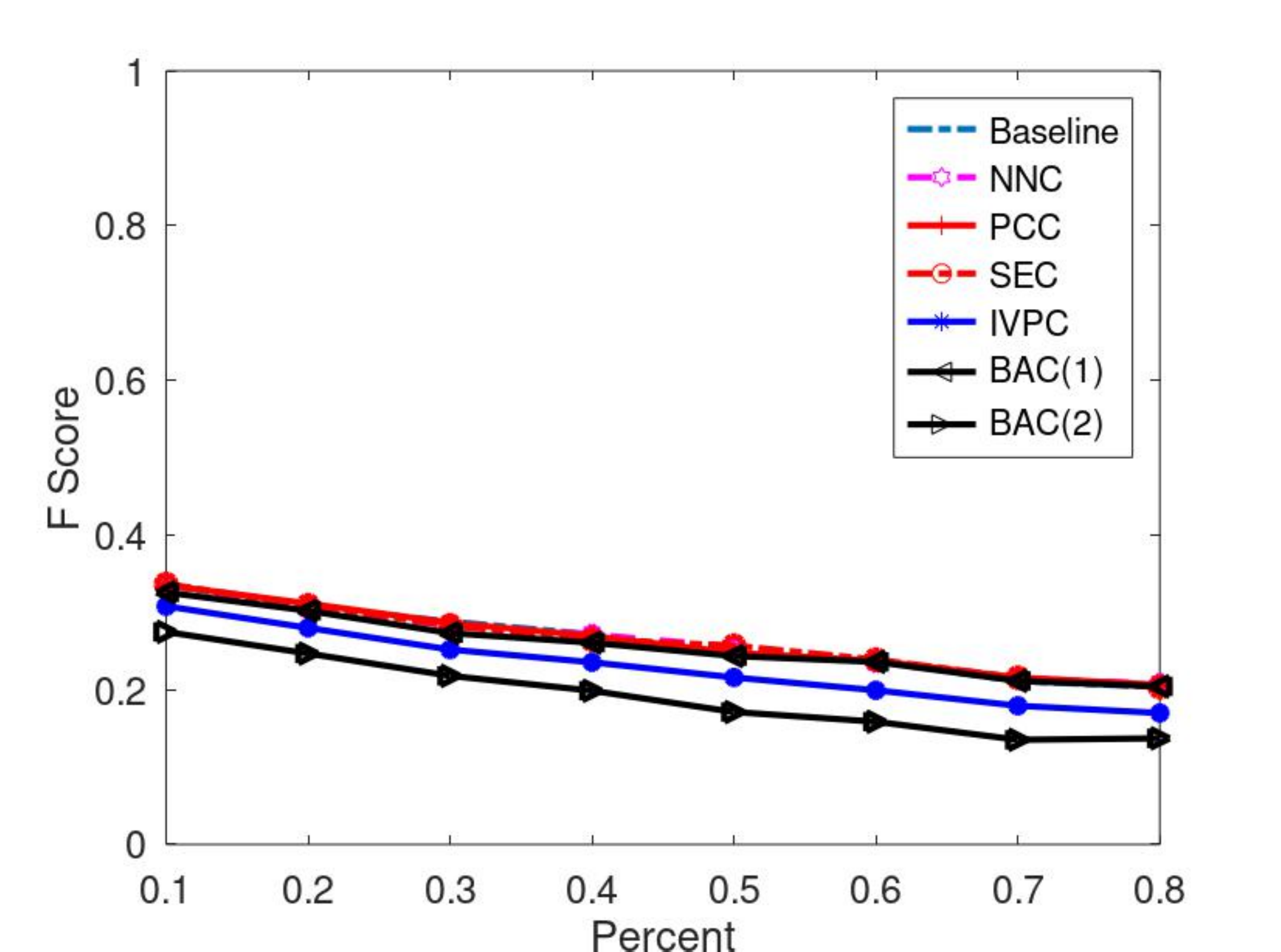}		}
    \subfigure[]{ \includegraphics[width=0.23\textwidth]{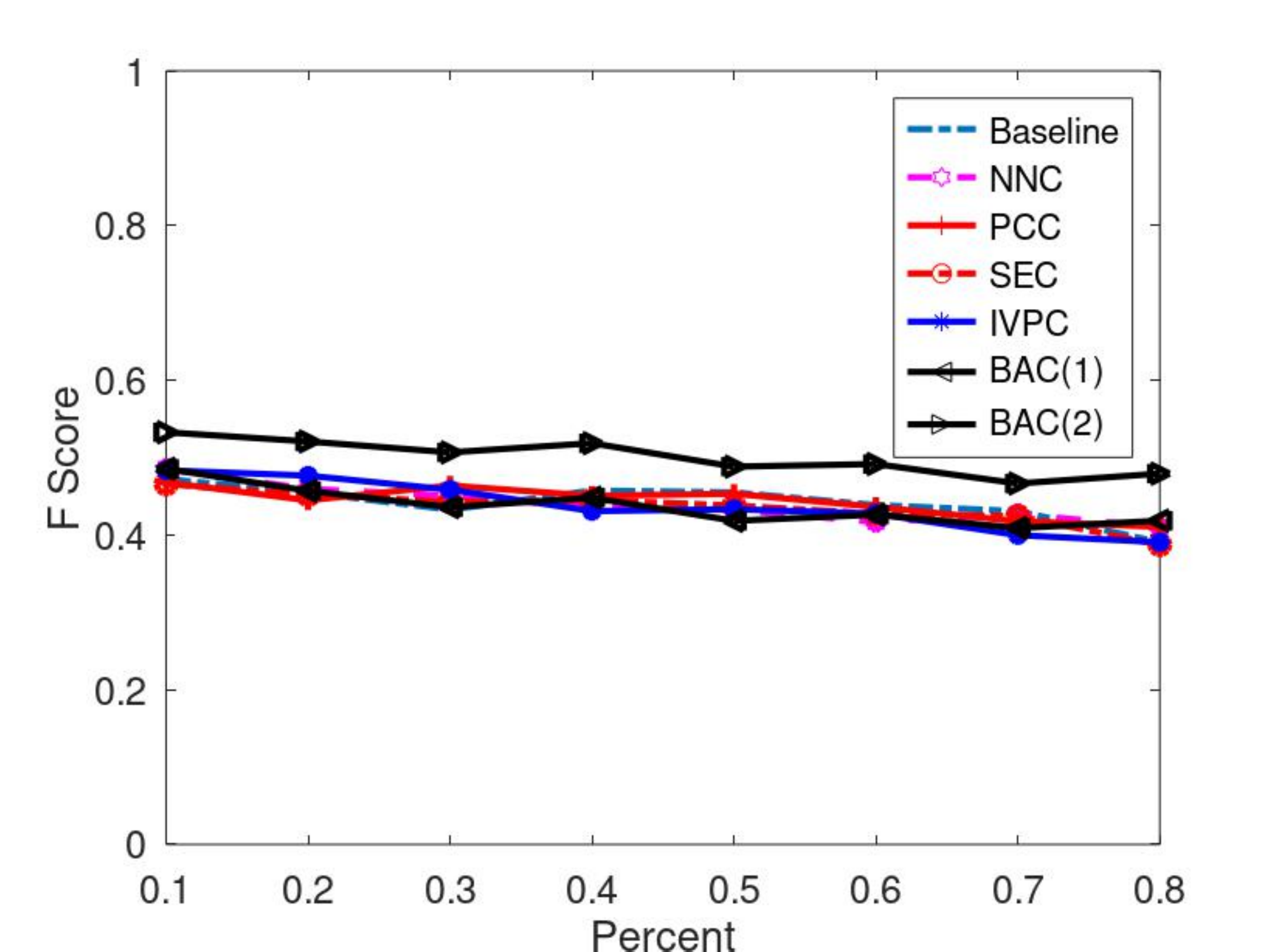}	}
    \subfigure[]{ \includegraphics[width=0.23\textwidth]{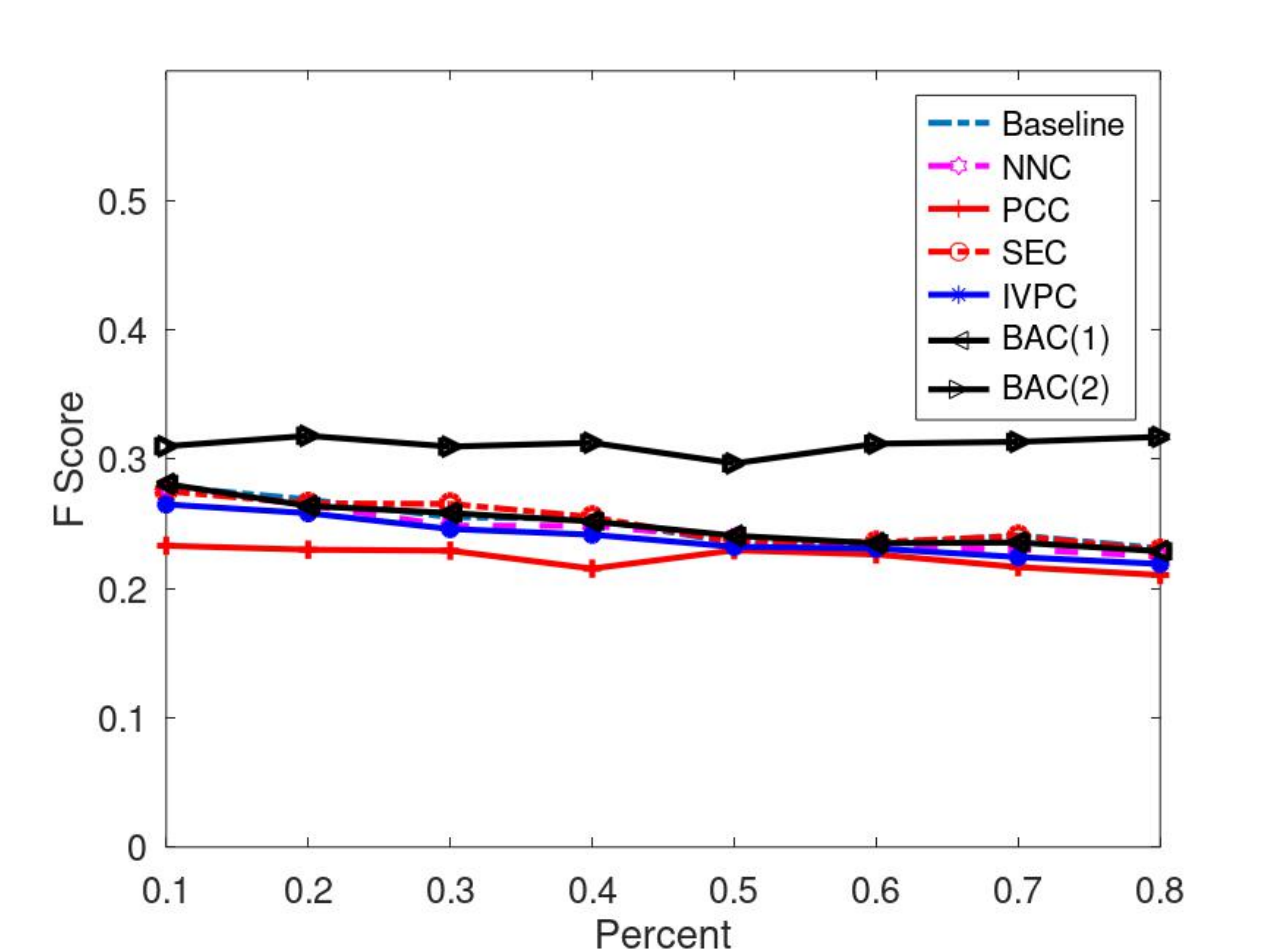}		}
    \subfigure[]{ \includegraphics[width=0.23\textwidth]{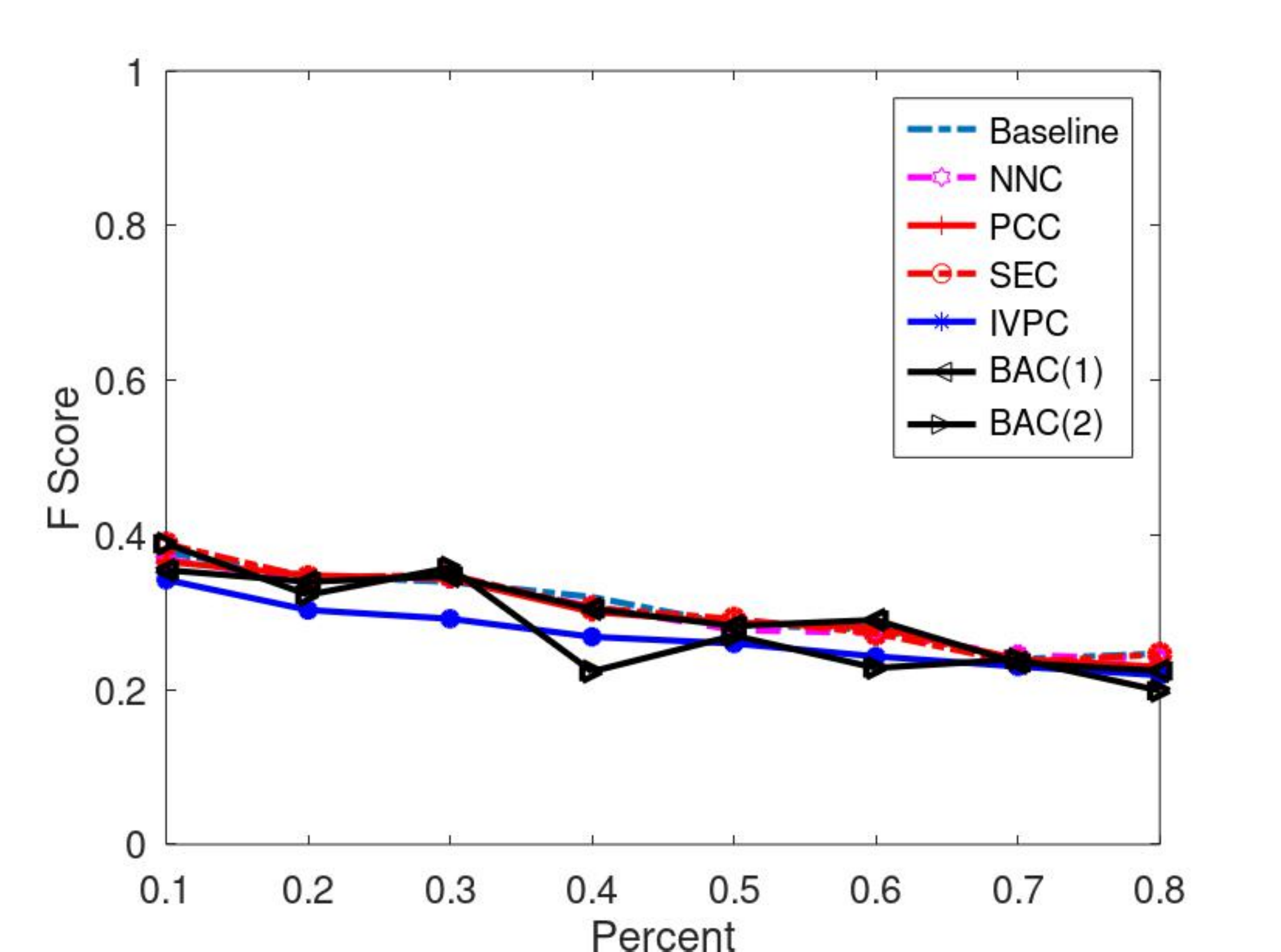}	}
    \caption{The obtained F Score of different algorithms associated with different percent of null elements on four data sets. (a) Birds (b) Firewall (c) Flower (d) Monkey.	 }
\end{figure*}
\begin{figure*}
    \centering
    \subfigure[]{ \includegraphics[width=0.23\textwidth]{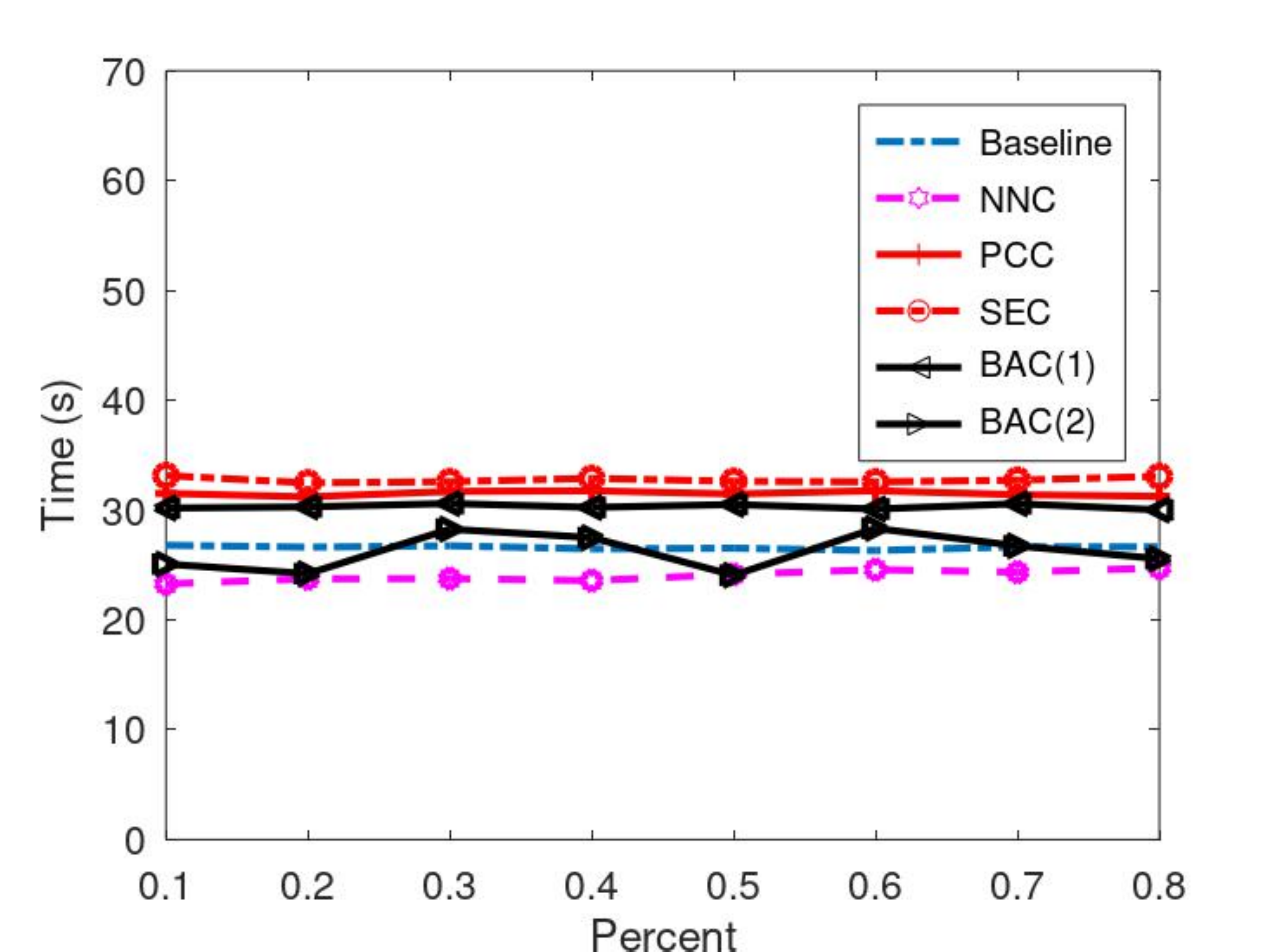}		}
    \subfigure[]{ \includegraphics[width=0.23\textwidth]{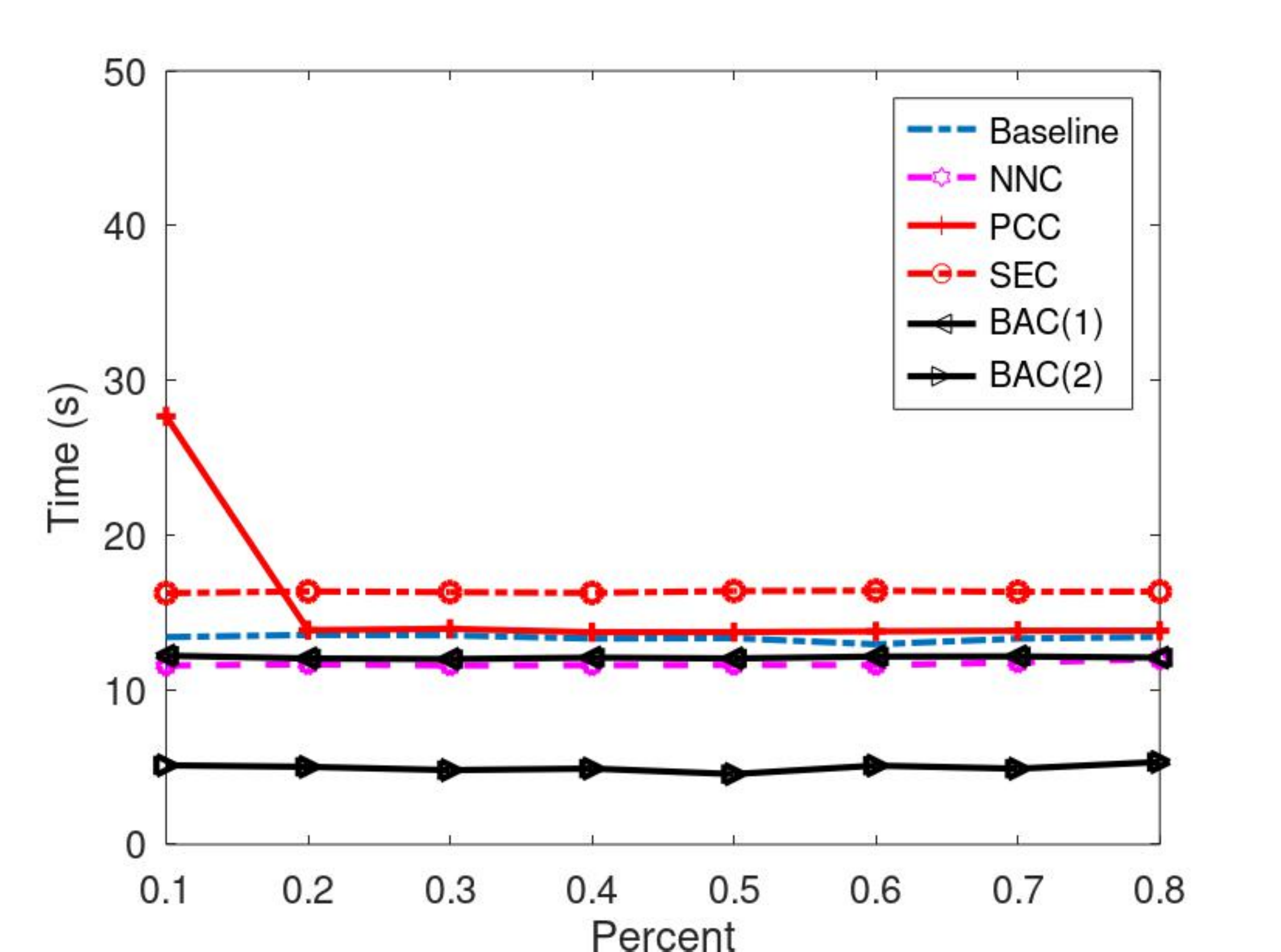}	}
    \subfigure[]{ \includegraphics[width=0.23\textwidth]{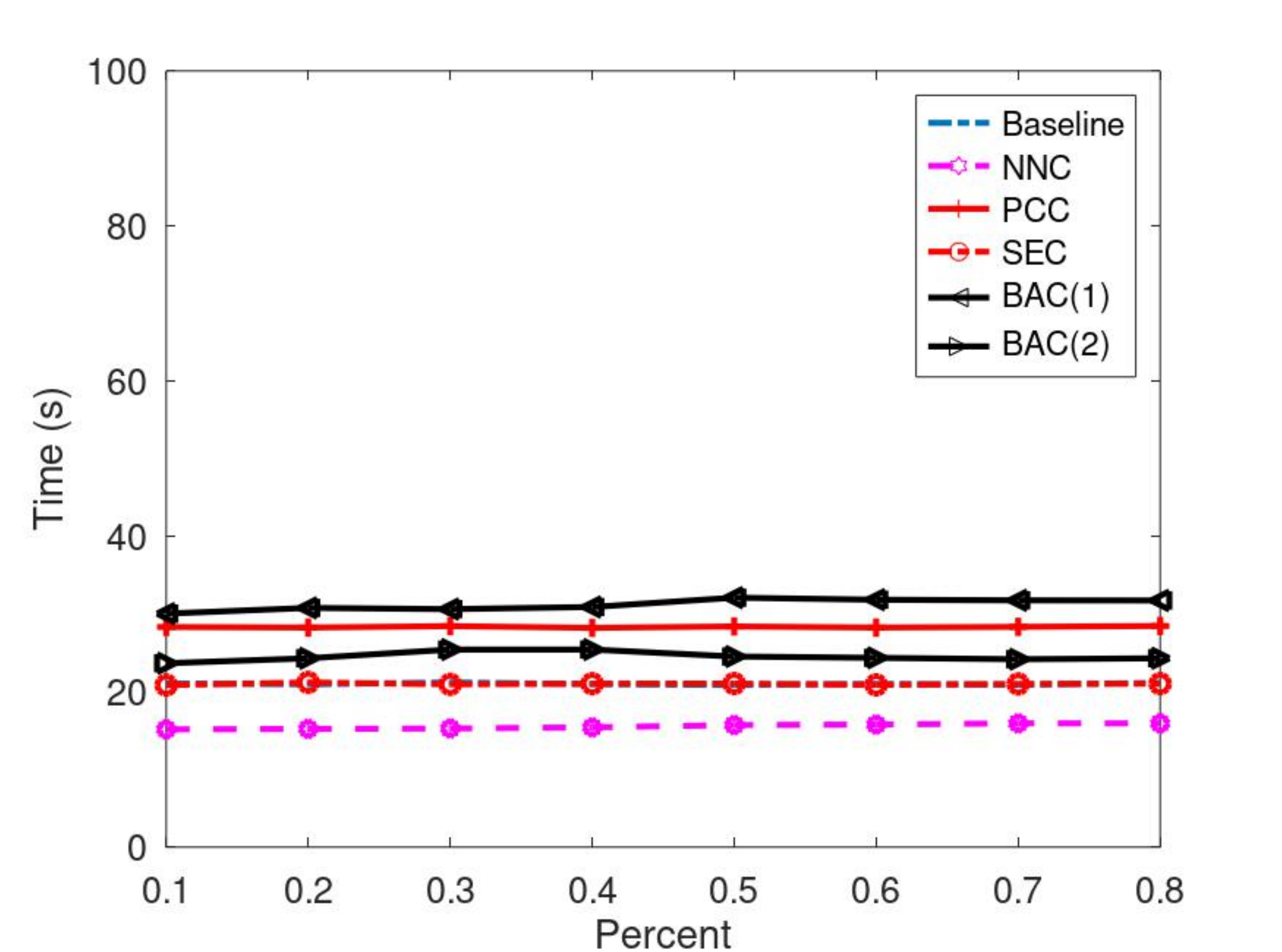}		}
    \subfigure[]{ \includegraphics[width=0.23\textwidth]{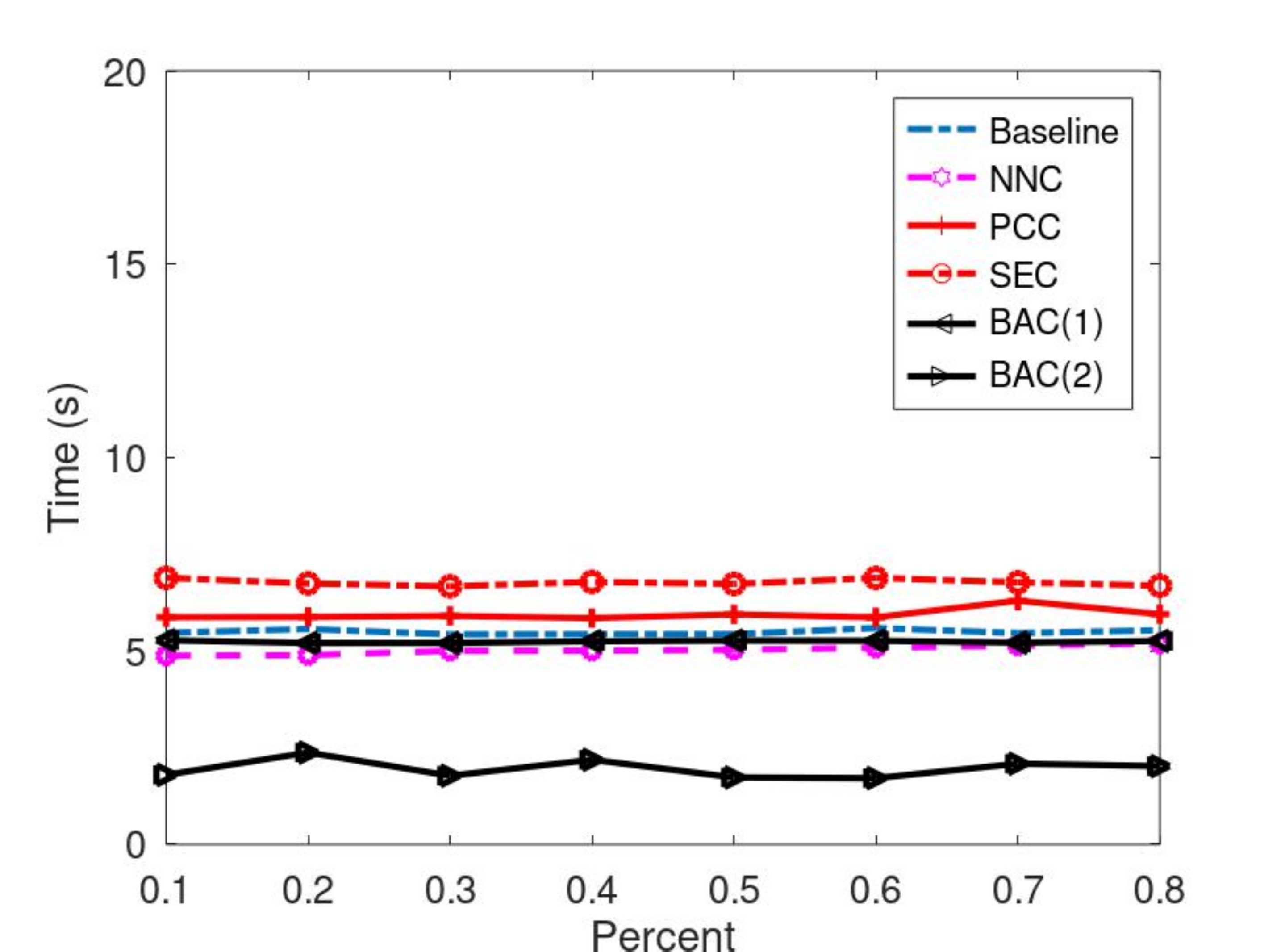}	}
    \caption{The time complexities (seconds) of different algorithms associated with different percent of null elements on four data sets. (a) Birds (b) Firewall (c) Flower (d) Monkey.	 }
\end{figure*}
In the second experiment, the performance of different methods with variational null elements are evaluated. More specifically, 0.1-0.8 percent of elements are randomly set to be null on four data sets, then all algorithms are repeatedly performed, and the average outputs are recorded as the results. In the experiment, the SEC based IVP method is performed to learn the data imputation associated with different percent of incomplete data. The NMI and F Score results are shown in Fig. 4-5, while Rand Index results are ignored due to the limited space.

In terms of the results, the performance of most algorithms are quite similar to each other, and the fluctuating tendency of average results are reduced slowly as more unavailable elements are appended into experiments. Obviously, all methods are affected by the involved amount of null elements, and the results of NNC, PCC, and SEC are quite close, owing to the fact that the final outlets depend on either imputation and clustering, especially on Birds and Firewall. Compared with SEC based BAC, \emph{k}-means BAC is superior than other methods, while IVPC is able to achieve comparable results.

In addition, the compuational costs associated with different percent of null elements are observed by recording the average time complexities of different methods, which is given in Fig. 6. Nevertheless, IVPC is ignored due to its high complexities of  calculation of information volumes for each instance.

In terms of the observations, the proposed BAC can present the ideal performance if \emph{k}-means is adopted to learn the partitions. Furthermore, stable time complexities are given by all methods as increasing parts of null elements, which discloses the efficiency of the involved methods. Since orthogonal decomposition is necessary for PCC, the most time complexities are required. Owing to random selection of neighbors, NNC method is able to present the optimistic efficiency on Birds and Flower data sets. Furthermore, SEC needs more computational times, as it is hardly to achieve stable convergence for self-expressive learning.

\section{Conclusion}
Clustering attempts to partition data instances into several groups associated with the maximum similarities of common characteristics. Furthermore, incomplete data frequently occurs in many real-world applications, e.g, digital data conveying and information processing, and brings perverse influence on data handling with missing values. In this work, a novel approach to clustering of incomplete data is proposed, which normally fulfill the incomplete data by leachable component learning.

More specifically, the proposed model exploits the similarities of distributions between the repaired and incomplete patterns, and alignment framework is adopted to learn the fulfilled elements. Similar to predication of Bayes distribution, it is able to afford clustering of incomplete data with estimation of distribution parameters.

Experiments on diverse artificial data sets show that, the proposed method is able to give the outstanding performance compared with the state-of-the-art methods, while calculation efficiency is still held. The future works mainly focus on the advances of the proposed method to other topics of information handling with incomplete data, as well as the relative learning of pattern analysis. 

%
\IEEEpeerreviewmaketitle

\section*{Acknowledgement}
The authors would like to thank anonymous reviewers for their constructive suggestions, and Firat University, General Dynamics, and Alexander Mamaev for providing data sets online. This work was partly supported by Innovation and Talent Foundation of Guangxi Province of China (RZ1900007485, AD19110154), and Natural Science Foundation of China (62172177). The corresponding author of this work is Miao Cheng.

\newpage



%
%
%

\bibliographystyle{IEEEtran}
\bibliography{lcc}

\end{document}